\pdfoutput=1

\documentclass[11pt]{article}

\usepackage[final]{acl}

\usepackage{latexsym}
\usepackage{times}
\usepackage{longtable}
\usepackage{subfigure}
\usepackage[subfigure]{tocloft}
\usepackage{xltabular}
\usepackage[T1]{fontenc}

\usepackage[utf8]{inputenc}

\usepackage{microtype}
\usepackage{amsmath}
\usepackage{hyperref}

\usepackage{inconsolata}
\usepackage{tocloft} 

\usepackage{graphicx}
\usepackage{bbding}

\usepackage{pifont}
\usepackage{tabularx} 
\usepackage{xcolor} 
\usepackage{wasysym}
\usepackage{booktabs}
\usepackage{subcaption} 
\usepackage{multirow} 
\usepackage{amssymb}

\definecolor{peach}{RGB}{223, 92, 94}
\definecolor{lightgreen}{RGB}{152, 251, 152}
\definecolor{darkgreen}{RGB}{0, 100, 0}

\title{PTD-SQL: Partitioning and Targeted Drilling with LLMs in Text-to-SQL}

\renewcommand{\thefootnote}{\fnsymbol{footnote}}
\author{
  Ruilin Luo$^{12}$$\footnotemark[1]$,~
  Liyuan Wang$^{2}$$\footnotemark[2]$,~
  Binghuai Lin$^{2}$,~ 
  Zicheng Lin$^{1}$,~
  Yujiu Yang$^{1}$$\footnotemark[2]$
  \\$^{1}$Tsinghua University
  \\$^{2}$Tencent Inc.
  \\ \texttt{lrl23@mails.tsinghua.edu.cn}
  \\ \texttt{sumerlywang@tencent.com,~yang.yujiu@sz.tsinghua.edu.cn}
}

\begin{document}
\maketitle
\begin{abstract}
Large Language Models (LLMs) have emerged as powerful tools for Text-to-SQL tasks, exhibiting remarkable reasoning capabilities. Different from tasks such as math word problems and commonsense reasoning, SQL solutions have a relatively fixed pattern. This facilitates the investigation of whether LLMs can benefit from categorical thinking, mirroring how humans acquire knowledge through inductive reasoning based on comparable examples. In this study, we propose that employing query group partitioning allows LLMs to focus on learning the thought processes specific to a single problem type, consequently enhancing their reasoning abilities across diverse difficulty levels and problem categories. Our experiments reveal that multiple advanced LLMs, when equipped with PTD-SQL, can either surpass or match previous state-of-the-art (SOTA) methods on the Spider and BIRD datasets. Intriguingly, models with varying initial performances have exhibited significant improvements, mainly at the boundary of their capabilities after targeted drilling, suggesting a parallel with human progress. Code is available at \href{https://github.com/lrlbbzl/PTD-SQL}{https://github.com/lrlbbzl/PTD-SQL}.
\end{abstract}
\footnotetext[1]{~Work done during Ruilin's internship at Tencent.}
\footnotetext[2]{~Corresponding author.}
\renewcommand{\thefootnote}{\arabic{footnote}}
\section{Introduction}

The Text-to-SQL task involves the automatic generation of SQL statements from natural language and has attracted much attention ~\cite{t2q_survey_1,qu2024before,jo2024lg}. Prior research primarily focused on training encoder-decoder models on text corpora and database schemas to capture generation patterns~\cite{xu2021sead}. Given the impressive capabilities of Large Language Models~(LLMs) in various Natural Language Processing~(NLP) tasks, numerous studies have endeavored to apply LLMs to this task~\cite{li2024can,zhang2024benchmarking,askari2024magic,lee2024mcs}.

\begin{figure}[!t]
    \centering
    \resizebox{\linewidth}{!}{
    \includegraphics[]{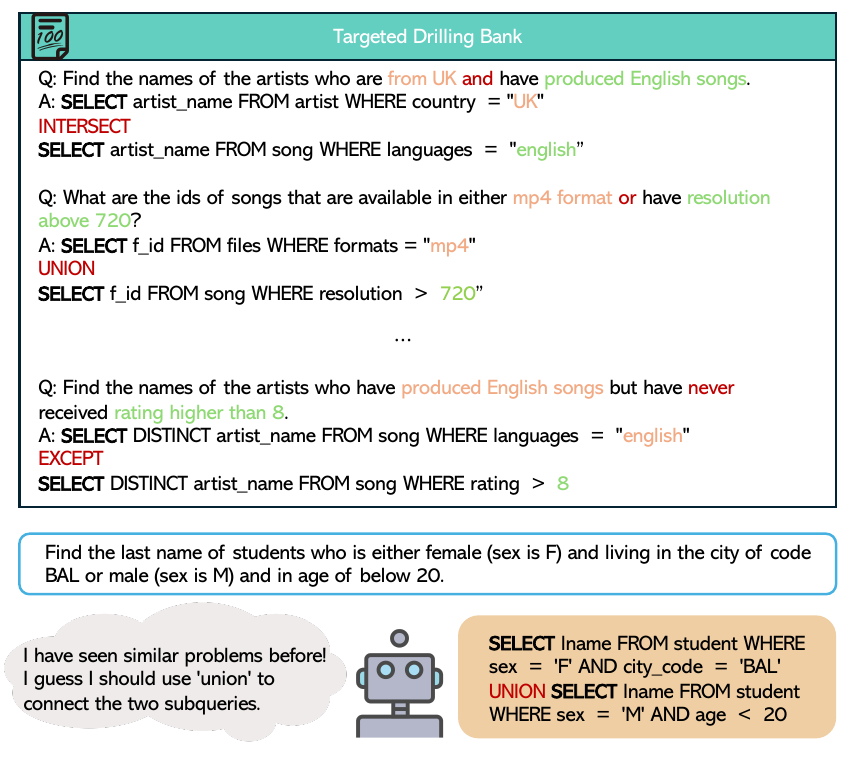}
    }
    \caption{Demonstration of targeted drilling prompt on multi-set problems.}
    \label{fig:teaser_graph}
\end{figure}

Recent investigations have proposed enhancing the reasoning capabilities of LLMs in the Text-to-SQL task, yielding substantial progress. Diverse methods such as the few-shot Chain-of-Thought~(CoT)~\cite{CoT}, self-consistency~\cite{self-consistency}, and the decomposition prompt that emphasizes dissecting complex problems and solving them sequentially~\cite{decompose1} have been introduced. A leading method, DIN-SQL~\cite{din-sql}, breaks down the task into several subtasks, classifies the complexity based on the nested logic of the problem, and applies different prompt strategies accordingly. However, like other studies, it overlooks the unique characteristics of SQL statements, which differ from math word problems and other code tasks. For calculations involving multiple sets, keywords like 'INTERSECT' or 'UNION' are often used to combine statements of several subproblems, making these queries naturally suitable for decomposition. Counting and sorting problems typically rely on 'GROUP BY' operations to identify objects to be aggregated and use 'ORDER BY' to sort other objects. Just like during a test with various question types, the knowledge points and problem-solving experiences that emerge in our minds are different.

Motivated by the brief overview of SQL question types above, we consider whether it is feasible to guide LLMs, akin to training human students for specific question types to master key concepts, by focusing on type-related examples during reasoning~\cite{self-discover}. Accordingly, we randomly select 100 multi-set operation questions from the training set, which require the use of keywords like 'INTERSECT' or 'EXCEPT'. We adopt two different prompt strategies: one from DIN-SQL, where these questions are classified as nested-level questions, providing samples of various question types under this complexity level, and another, as depicted in Figure~\ref{fig:teaser_graph}, where we only provide LLM multi-set question examples with the same number. With these strategies, we achieve execution accuracy rates of 39.0 and 55.0 using ChatGPT, respectively. The former exhibits more sub-query errors and logical confusion.

Drawing on the above observation, we propose the \textbf{P}artitioning and \textbf{T}argeted \textbf{D}rilling (PTD-SQL) framework to enhance LLMs' reasoning capabilities in Text-to-SQL tasks. This strategy mirrors the human learning process, where students typically first identify the group of questions and then search for the most relevant knowledge points to answer them. Initially, we categorize the types of textual queries in the training set based on the keywords in the ground-truth SQL statements. Informed by previous studies, we opt not to rely solely on the LLM's few-shot discrimination ability but instead delegate a small LLM with fine-tuning for this task~\cite{daslam,zhuang2023efficiently}. In the second step, we design prompts with different emphases for various categories of problems in the training set and automatically generate problem sets and reference answers – the areas that the LLM needs to learn. Both of these operations are performed offline and avoid invoking GPT during testing, thus achieving cost efficiency. Finally, during the inference stage, we classify the original textual query and design an automatic selection module to compose a few-shot prompt in the corresponding group of the problem set~\cite{an2023skill}. 

We extensively validate the effectiveness of PTD-SQL on the Spider-dev, Spider-realistic, and BIRD-dev datasets using three powerful LLMs, where it outperforms state-of-the-art frameworks such as DIN-SQL and DAIL-SQL. We also find that the model becomes more capable of achieving breakthroughs at the capability boundaries when equipped with PTD-SQL, which may potentially extend to other reasoning tasks. Furthermore, our approach adheres to a one-time query paradigm, showing advantages in terms of token consumption and inference time, also allowing many methods targeting schema linking or database content alignment to be seamlessly integrated, thereby anticipating even higher performance.

\begin{figure}[!t]
    \centering
    \resizebox{\linewidth}{!}{
    \includegraphics[]{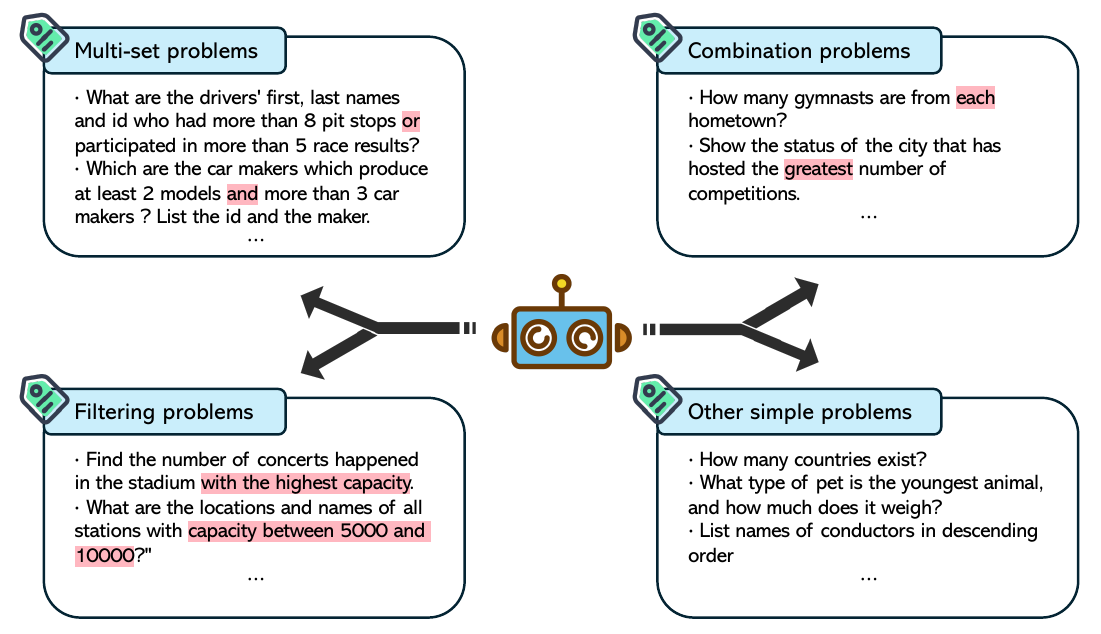}
    }
    \caption{Some samples of proposed partition.}
    \label{fig:groups}
\end{figure}

\section{Related Work}
\paragraph{LLM Reasoning}
Nowadays, the development of reasoning models based on LLM has become a popular and critical field. Many efficient prompting methods have been proposed, such as Chain-of-Thought~\cite{CoT}, which guides LLM in step-by-step thinking; Least-to-Most~\cite{ltm}, which makes the model adapt to the difficulty gradient; and Decomposition-based prompting~\cite{decompose1,decompose2}, which breaks down difficult problems to solve them separately. In addition, Self-Consistency~\cite{self-consistency} demonstrates the overall tendency of LLM towards the correct answer through voting, Self-discover~\cite{self-discover} allows the model to make different problem-solving plans according to different types of questions, and Self-refine~\cite{self-refine} enables LLM to learn from the feedback of its problem-solving process. Besides, many works also strengthen the weaker aspects of LLM at the code level, such as PAL~\cite{PAL} and PoT~\cite{PoT}. 

\begin{figure*}[!t]
    \centering
    \resizebox{\linewidth}{!}{
    \includegraphics[]{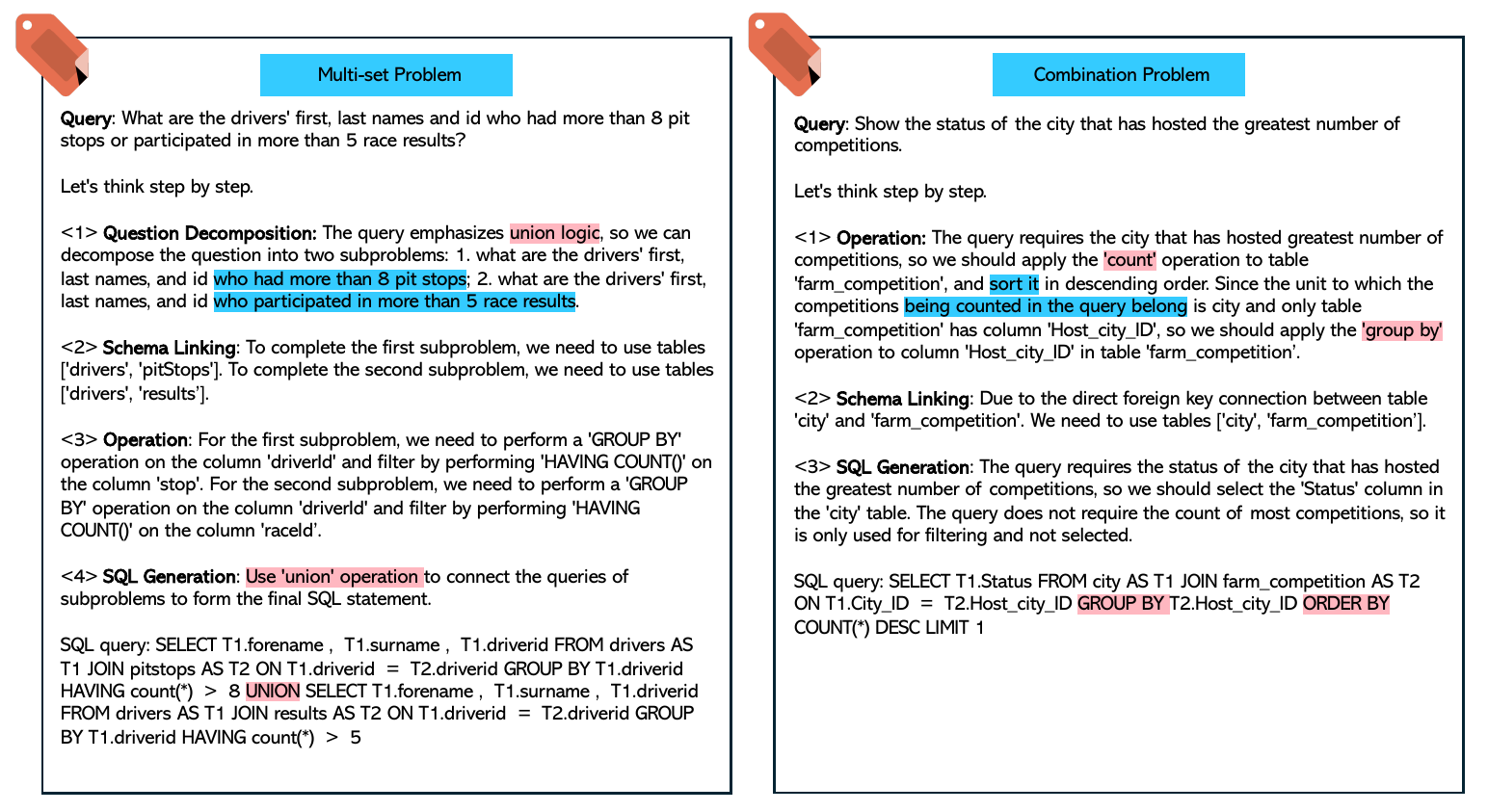}
    }
    \caption{Prompt demonstrations for Multi-set and Combination problem.}
    \label{fig:construction_prompt}
\end{figure*}

\paragraph{LLM-based Text-to-SQL}
Nowadays, many studies are focusing on utilizing LLMs to complete Text-to-SQL tasks, primarily involving more efficient prompt design and advanced process deployment. Strategies that have proven effective in common sense reasoning and mathematical reasoning, such as CoT and self-consistency, have also been applied to enhance Text-to-SQL reasoning. C3~\cite{c3-sql} and StructGPT~\cite{structgpt} have introduced effective zero-shot strategies based on GPT, along with meticulous interface settings. DIN-SQL~\cite{din-sql} divides the Text-to-SQL task into phased subtasks and assigns different LLMs to specialize in completing each stage, as well as categorizes the difficulty of questions to provide varying prompt strategies. DAIL-SQL~\cite{dail-sql} has conducted a comprehensive evaluation of many prompt-based methods and proposed a more precise samples matching approach to improve results. Recent approaches have also concentrated on addressing issues not yet considered in the data itself. For instance, PET-SQL~\cite{pet-sql} focuses on leveraging prior knowledge within databases to enhance the accuracy of responses at the token level, which shows benefit on Text-to-SQL. SQL-CRAFT~\cite{sql-craft} suggests allowing models to engage in interactive refinement to improve reasoning accuracy. DEA-SQL~\cite{xie2024decomposition} and MAC-SQL~\cite{wang2023mac} integrate multiple optimization techniques to propose workflow agents. Recently, many new benchmarks have been proposed for the development of this field to accommodate more enterprise-level applications~\cite{saparina2024ambrosia,zhou2024db}.

\section{Pipeline of PTD-SQL}
\begin{figure*}[!t]
    \centering
    \resizebox{\linewidth}{!}{
    \includegraphics[]{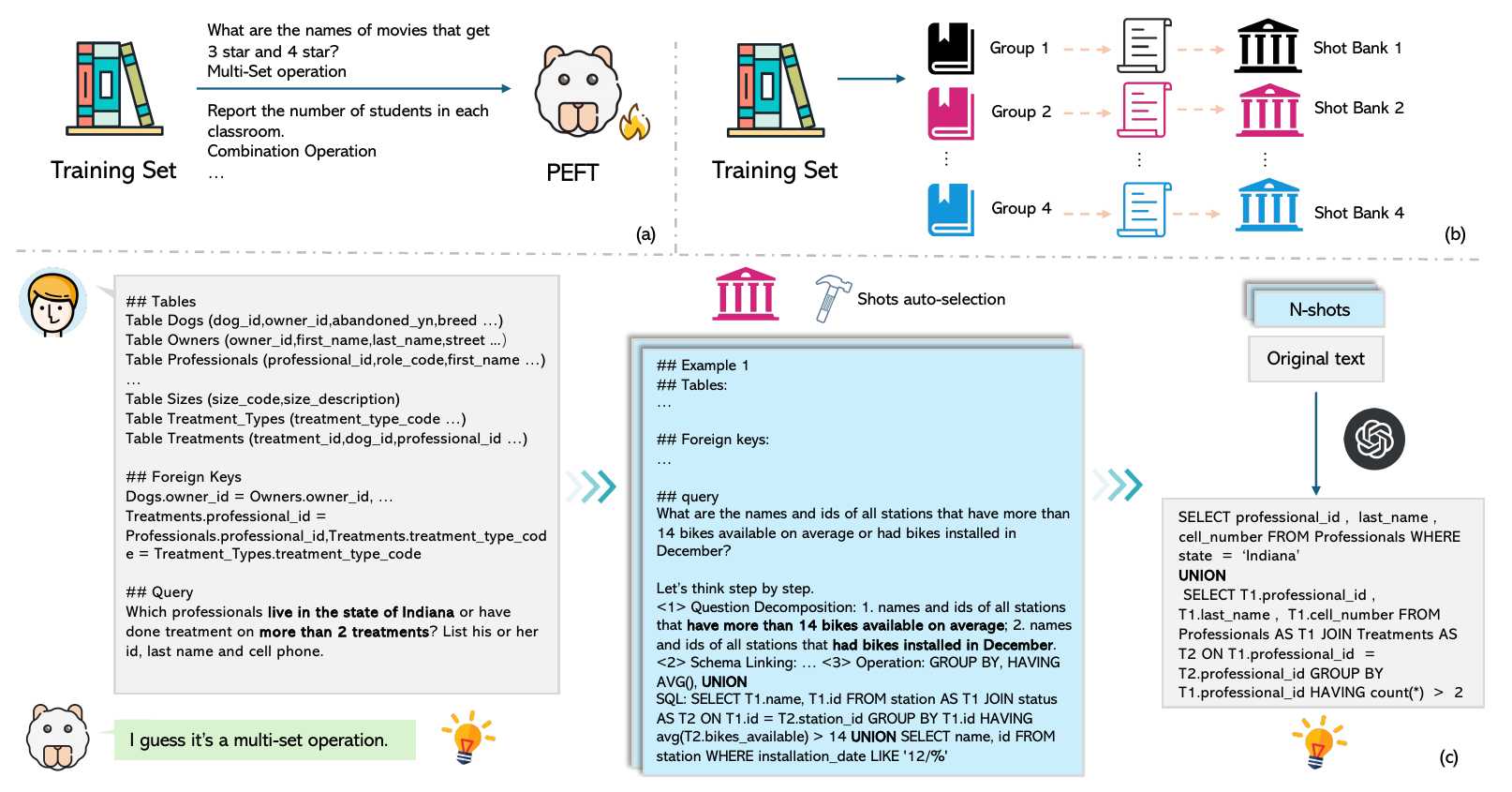}
    }
    \caption{Overflow of PTD-SQL. (a) QGP sub-task. (b) Targeted drilling bank auto-construction. (c) Reasoning step.}
    \label{fig:overview}
\end{figure*}

In this section, we present the process of the PTD-SQL framework as illustrated in Figure~\ref{fig:overview}, which includes: \romannumeral1. The design and implementation of the proposed Query Group Partition~(QGP) sub-task; \romannumeral2. The automatic construction of distinct query group question banks, each containing its unique reasoning process; \romannumeral3. The inference process.

\subsection{Query Group Partition}\label{sec:qgp}
In this section, we first provide the definition of the QGP sub-task and then describe the process of fine-tuning the small LLM using PEFT to accomplish the QGP task.

\paragraph{Problem Formulation}
SQL queries differ from math word problems and other code problems, such as Python, as their textual labels often contain highly characteristic expressions, making problem group identification convenient. We cluster them based on label keywords: multi-set, combination, filter, and other simple problems. Multi-set problems frequently involve two or more layers of logic and require keywords like 'INTERSECT', 'UNION', and 'EXCEPT' for connection. Combination problems necessitate the use of a 'GROUP BY' operation to group data, followed by sorting, taking extreme values, and other purposeful operations. Filter problems involve constructing conditional statements and using them for target screening. The remaining problems are classified as other simple problems, as depicted in Figure~\ref{fig:groups}. Considering that some queries may have implicit labels of other types, we provide prioritized classification criteria in the prompt to alleviate the impact of model bias. Specific examples are shown in the Appendix~\ref{app:sec:qgp_prompt}. The task objective is explicitly defined as follows: given a text query $q$, we need to output its problem group $\hat g$. It is formulated as:

\begin{equation}
    \begin{aligned}
        \hat g = f(q\ |\ \theta)
    \end{aligned}
\end{equation}
where $f(\cdot\ |\ \theta)$ can present a model with parameters $\theta$. We randomly select the training set $\mathcal{S}_{T}$ for the QGP task on the original training set and separate the validation set $\mathcal{S}_V$.

\paragraph{Fine-tuned LLM Classifier}
Inspired by previous works~\cite{daslam,zhuang2023efficiently}, we consider delegating the ability to determine categories to the fine-tuning process of the small LLM rather than directly trusting the discrimination capability of LLM. With the rapid advancement of PEFT technology, we choose Low-Rank Adaptation~(LoRA)~\cite{lora} to fine-tune the Llama-2-7b model to solve the QGP problem. For a pre-trained weight matrix $W_0\in R^{d\times k}$, LoRA adds a bypass using two decomposition matrices $A\in R^{d\times r}$ and $B\in R^{r\times k}$, where $r \ll min(d,k)$. The forward process of single weight matrix is modified to:

\begin{equation}
    \begin{aligned}
        h=W_0x+BAx
    \end{aligned}
\end{equation}
During finetuning with LoRA, we freeze the original weights of LLM and only update low-rank matrices $A$ and $B$. 

For annotated labels $G$ and outputs of LLM, the objective loss is defined as :
\begin{equation}
    \begin{aligned}
        \mathcal{L}=CrossEntropy(G\ |\ f(q\ |\ \theta + \delta \theta))
    \end{aligned}
\end{equation}

\begin{table}[!t]
\centering
\begin{tabular}{l|ccc}
\hline
\textbf{Method} & \textbf{Exact Match}\\
\hline
Llama-2-7b\ +\ LoRA & 85.0\%\\
ChatGPT\ +\ 10-shot & 68.0\%\\
\hline
\end{tabular}
\caption{Performance on validation set of QGP sub-task.}
\label{tab:qgp_result}
\vspace{-0.2in}
\end{table}

\paragraph{Finetuned Small Model vs. Few-shot GPT} 
The performance of the fine-tuned Llama-2-7b model and the few-shot prompting ChatGPT on the QGP task is presented in Table~\ref{tab:qgp_result}. This highlights the superiority of PEFT in downstream tasks and prompts us to use the former on the test set.

\subsection{Targeted Drilling Bank Auto-construction}
\label{sec:auto-construction}
In this section, we explain how to construct targeted drilling banks for different question groups in PTD-SQL, which can be compared to the specialized training and reference ideas and answers designed by teachers for students before examinations. Previous works grade the difficulty based on whether the problem requires nesting and designing corresponding prompt templates. However, this approach only focuses on the surface logic of SQL queries and does not consider the distinct thinking paths required by the essence of different question groups for LLM. Given that selecting irrelevant examples may also be detrimental to LLM's thinking, in PTD-SQL, we can benefit from the proposed QGP. That is, for test queries of specific question groups, we can directly and accurately locate the problem banks with similar thinking paths.

Multi-set problems often require breaking down a complex problem into multiple subqueries and integrating the different results through connecting keywords. For filtering problems, we can often prompt LLM to first propose the organization of filtering conditions and then process the selection target. Therefore, these two types of problems are naturally suitable for the design inspiration of decomposed prompting. We show an example of prompt construction for a multi-set problem, as depicted in Figure~\ref{fig:construction_prompt}. For filtering problems, our decomposition focuses on the division of conditional statements and the extraction of target columns, and the specific prompts are shown in Appendix~\ref{sec:app_bank_prompt}. It is worth mentioning that we treat schema linking as a byproduct of LLM's thinking process, thereby achieving the purpose of one-time generation, which reduces the query cost.

For combination problems and other simple problems, we construct concise CoT templates. For the former, the model is required to distinguish the objects that need to be counted (sorted or taking extreme values) and the groups they belong to, thus improving the ability to organize answers under this question type. An example is shown in Figure~\ref{fig:construction_prompt}. For the remaining simple problems, we choose to use the ground truth SQL query directly as the composition of the few-shot prompt without introducing other thinking processes.

After creating four different types of few-shot prompts, we apply them separately to their respective problem groups in the training set to generate the thinking process and the final SQL query. We select the samples with correct execution results of the SQL query to form four targeted drilling banks because we believe that the thinking paths in the examples with correct final answers are highly likely to be reasonable and enlightening. These are the sources of the examples that LLM refers to during the inference phase. The specific statistics of different targeted drilling banks are shown in Appendix~\ref{app:sec:bank_statistics}.

\subsection{Few-shot Selection}
Few-shot example construction is a crucial step in prompt engineering because LLMs are sensitive to few-shot samples. In PTD-SQL, we perform QGP on each textual query and then automatically select shots in the corresponding targeted drilling bank.

\paragraph{Semantic matching} Previous work has verified the effectiveness of methods based on semantic vector matching~\cite{an-etal-2023-skill}. We calculate and store sentence embeddings for all textual queries in the targeted drilling bank using OpenAI text-embedding-ada-002\footnote{\href{https://platform.openai.com/docs/guides/embeddings}{https://platform.openai.com/docs/guides/embeddings}}, resulting in an offline bank matrix $\mathcal{M}$. For test queries, we encode them with text-embedding-ada-002 and calculate the cosine similarity with $\mathcal{M}$ to measure the degree of semantic matching as some previous works do. 
\begin{equation}
\resizebox{0.8\hsize}{!}{
        $sim_1(s, s_i) = \frac{Emb(s)Emb(s_i)^T}{|Emb(s)||Emb(s_i)|}$}
\end{equation}

\paragraph{Syntactic matching} Considering that textual SQL queries have strong syntactic features, such as counting problems often having phrases like "how many", and extreme value demands often accompanied by comparative adjectives like "largest" or "lowest". Therefore, we use token overlap counts to rank the syntactic relevance of samples in the corresponding targeted drilling bank. 

\begin{equation}
\resizebox{\hsize}{!}{
        $sim_2(s,s_i) = \frac{len(set(tokenize(s))\ \&\ set(tokenize(s_i)))}{len(set(tokenize(s)))}$
    }
\end{equation}

\paragraph{Mix-of-matching} Similar to the idea of multi-way recall, we mix an equal amount of examples selected by the two strategies above, for instance, choosing the top 2 most relevant examples from each in a 4-shot scenario, in order to provide as rich and relevant samples as possible within the same problem group, thus guiding effective thinking.

\section{Experiments}
\subsection{Experimental Setup}
\label{sec:exp}
\paragraph{Datasets} Spider~\cite{spider} is the most widely used cross-domain dataset. This dataset has 7,000 training data in the training set and 1,034 data in the development set, covering 200 different databases and spanning 138 domains. Spider-realistic~\cite{spider_real} is a more challenging dataset containing 508 test data points, which manually mask the specific column selections in the text query.  BIRD~\cite{li2024can} dataset contains 95 large-scale real databases covering 37 professional domains. More details and usage of the data can be found in Appendix~\ref{app:data_statistics}.

\paragraph{Evaluation} Most previous work adheres to two common evaluation metrics: 1) Exact Match Accuracy (EM): It requires that each subcomponent of the SQL query generated by the model matches the gold SQL query provided in the dataset. 2) Execution Accuracy (EX): EX judges correctness based on whether the answer returned by executing the predicted SQL query in the database is consistent with the gold query. Since a textual query may correspond to several correct but stylistically different SQL query formulations, it is a more accurate measure of Text-to-SQL methods. Besides, Valid Efficiency Score~(VES) is used to demonstrate the efficiency of valid SQLs provided by models.

\begin{table}[!t]
\centering
\resizebox{\linewidth}{!}{
\begin{tabular}{l|c|c}
\hline
Methods & Type & EX\\
\hline 
T5-3B + PICARD$^\dagger$~\cite{picard} & Fine-tuning & 79.3 \\ 
RESDSQL + NatSQL$^\dagger$~\cite{li2023resdsql} & Fine-tuning &  \underline{84.1} \\
C3 + ChatGPT$^\dagger$~\cite{c3-sql} & Zero-shot & 81.2 \\
ChatGPT~\cite{comprehensive} & Zero-shot & 70.1 \\
GPT-4~\cite{gpt4,dail-sql} & Zero-shot & 72.3 \\
\midrule
DIN-SQL + ChatGPT$^\S$~\cite{din-sql} & Few-shot & 76.8 \\
DIN-SQL + GPT-4$^\S$ & Few-shot & 80.6 \\
DIN-SQL + {Deepseek-coder-6.7b-instruct}$^\ddagger$ & Few-shot & 73.6 \\
\midrule
DAIL-SQL + ChatGPT$^\dagger$~\cite{dail-sql} & Few-shot & 79.1 \\
DAIL-SQL + GPT-4$^\dagger$ & Few-shot & 83.1 \\
DAIL-SQL + GPT-4 + Self-Consistency$^\dagger$ & Few-shot & 83.6 \\ 
DAIL-SQL + Deepseek-coder-6.7b-instruct$^\ddagger$ & Few-shot & 75.7 \\
\midrule
PTD-SQL + ChatGPT$_{ours}$ & Few-shot & 80.3 \\
PTD-SQL + GPT-4$_{ours}$ & Few-shot & \textbf{85.7} \\
PTD-SQL + Deepseek-coder-6.7b-instruct$_{ours}$ & Few-shot & 76.7 \\ 

\bottomrule
\end{tabular}
}
\caption{EX on Spider-dev set. Results of methods with $\dagger$ are taken from the original paper or open-source code repository. Results with label $\ddagger$ are implemented by us. Results with $\S$ are obtained from the running results files provided by~\cite{din-sql} and evaluation program~\cite{eval}.}
\label{tab:main_result}
\end{table}

\paragraph{Baselines} We compare three different path Text-to-SQL methods, including fine-tuning, zero-shot, and few-shot prompting methods. Among them, the fine-tuning method includes PICARD~\cite{picard} and the current SOTA RESDSQL+NatSQL~\cite{li2023resdsql}. The zero-shot method C3~\cite{c3-sql} focuses on schema linking filtering and removing GPT's inherent bias for SQL generation. DIN-SQL~\cite{din-sql}, which breaks down the textual query into multiple staged questions. DAIL-SQL~\cite{dail-sql} considers optimizing sample selection and organization to further enhance LLM's reasoning ability in Text-to-SQL.

\paragraph{Implementation Details} In order to comprehensively evaluate the performance of the framework on closed-source and open-source models and demonstrate its effectiveness, we employ three LLMs for comparison purposes: OpenAI GPT-3.5-turbo-0613 for ChatGPT, GPT-4-0613, and Deepseek-coder-6.7b-instruct\footnote{\href{https://huggingface.co/deepseek-ai/deepseek-coder-6.7b-instruct}{https://huggingface.co/deepseek-ai/deepseek-coder-6.7b-instruct}}~\cite{guo2024deepseek}. The latter is pretrained on high-quality code corpora and has attained the current state-of-the-art performance among open-source code models in the realm of code generation. Maximum context length is limited to 4096 for OpenAI LLMs and 2048 for open-source LLMs.

\begin{table}[!t]
\centering
\resizebox{\linewidth}{!}{
\begin{tabular}{l|c|c}
\hline
Methods & Type & EX\\
\hline 
ChatGPT & Zero-shot & 67.3  \\ 
GPT-4 & Zero-shot & 66.5 \\
\midrule
DIN-SQL+ChatGPT & Few-shot & \underline{70.3} \\
DIN-SQL+Deepseek-coder-6.7b-instruct & Few-shot & 68.3 \\
\midrule
DAIL-SQL+ChatGPT & Few-shot & 69.3 \\
DAIL-SQL+ Deepseek-coder-6.7b-instruct & Few-shot & 68.9 \\
\midrule
PTD-SQL+ChatGPT$_{Ours}$ & Few-shot & \textbf{72.2} \\
PTD-SQL+Deepseek-coder-6.7b-instruct$_{ours}$ & Few-shot & 69.9 \\
\hline
\end{tabular}}
\caption{EX on Spider-realistic dataset.}
\vspace*{-2mm}
\label{tab:realistic_result}
\end{table}

\begin{table}[!t]
\centering
\resizebox{\linewidth}{!}{
\begin{tabular}{l|c|c}
\hline
Methods  & EX & VES\\
\hline 
CodeX & 34.4 & 41.6 \\
ChatGPT+CoT & 36.6 &42.3 \\
GPT-4 & 46.4 & 49.8\\
\midrule
DIN-SQL + ChatGPT & 41.0 & 51.4\\
DIN-SQL + GPT-4 & 50.2 & \textbf{58.1}\\
DIN-SQL + {Deepseek-coder-6.7b-instruct} & 40.7 & 49.0\\
\midrule
DAIL-SQL + ChatGPT & 41.2 & 49.2 \\
DAIL-SQL + GPT-4 & \underline{53.6} & 56.5 \\
DAIL-SQL + Deepseek-coder-6.7b-instruct & 42.4 & 50.2 \\
\midrule
PTD-SQL + ChatGPT$_{ours}$  & 44.2 & 53.3\\
PTD-SQL + GPT-4$_{ours}$  & \textbf{57.0} & \underline{57.7} \\
PTD-SQL + Deepseek-coder-6.7b-instruct$_{ours}$ & 45.4 & 55.0\\ 

\hline
\end{tabular}
}
\caption{EX and VES comparison on BIRD dataset.}
\label{tab:bird_result}
\end{table}

\subsection{Main Results}
As shown in Table~\ref{tab:main_result}, PTD-SQL + GPT4 achieves the best EX metric on the Spider-dev dataset. Additionally, PTD-SQL surpasses DIN-SQL and DAIL-SQL when using ChatGPT and Deepseek-coder-6.7b-instruct as base models. Compared to the more advanced DAIL-SQL framework, PTD-SQL achieves relative increases of 1.5\%, 3.1\%, and 1.3\% on ChatGPT, GPT-4 and Deepseek-coder-6.7b-instruct, respectively. When compared with previous fine-tuning and prompting methods, PTD-SQL also attains a comparative performance. Besides, as shown in Table~\ref{tab:realistic_result}, ChatGPT-equipped PTD-SQL also outperforms previous methods and GPT-4 using zero-shot. Furthermore, the results shown in Table~\ref{tab:bird_result} indicate that all three powerful models equipped with PTD-SQL demonstrate stronger EX. In terms of VES indicators, PTD-SQL also has a certain competitiveness. A case study on Spider is given in Appendix~\ref{app:sec:case_study}. Furthermore, we discuss the advantages of PTD-SQL in terms of token consumption and inference time in Appendix~\ref{app:sec:time_token_cost}.


\begin{table}[!t]
    \centering
    \resizebox{0.9\linewidth}{!}{
    \begin{tabular}{c|c|c|c|c|c|c}
      \toprule
      Base Model & Method &  Easy & Medium & Hard & Extra & All \\
      \midrule
      \multirow{3}{*}{\shortstack{Deepseek-coder\\-6.7b-instruct}} & DIN-SQL & 86.3 & 81.2 & 59.8 & 48.8 & 73.6 \\
             & DAIL-SQL & 86.7 & \textbf{81.6} & 59.2 & 50.0 & 75.7 \\
             & PTD-SQL & \textbf{87.1} & 78.9 & \textbf{74.9} & \textbf{57.2} & \textbf{76.7} \\
        \midrule
        \multirow{3}{*}{ChatGPT} & DIN-SQL & 90.7 & 82.3 & 62.1 & \textbf{56.6} & 76.8 \\
             & DAIL-SQL & \textbf{91.5} & \textbf{83.8} & 71.2 & 56.0 & 79.1 \\
             & PTD-SQL & 90.7 & 83.1 & \textbf{80.6} & \textbf{56.6} & \textbf{80.3} \\
        \midrule
        \multirow{3}{*}{GPT-4} & DIN-SQL & 89.9 & 84.3 & 78.2 & 57.8 & 80.4 \\
             & DAIL-SQL & 90.7 & \textbf{89.7} & 75.3 & 62.0 & 83.1 \\
             & PTD-SQL & \textbf{94.8} & 88.8 & \textbf{85.1} & \textbf{64.5} & \textbf{85.7} \\
    
      \bottomrule
    \end{tabular}
    }
    \caption{Performance comparison on three LLMs across difficulty levels on Spider-Dev dataset.}
    \label{tab:level}
  \end{table}

\section{More Discussion}
In this section, we investigate the efficacy of PTD-SQL, taking into account both the challenges posed by the database itself (\textbf{RQ1}) and the performance across various problem groups~(\textbf{RQ2}). Concurrently, we delve into the insights that PTD-SQL contributes to the LLM-based Text-to-SQL domain. Furthermore, we perform ablation studies on the employed modules, primarily focusing on the effectiveness of introduced QGP task~(\textbf{RQ3}), and the influence of shot selection strategies within the same targeted drilling bank (\textbf{RQ4}).

\subsection{RQ1: Performance from a Difficulty-level}

In this subsection, we evaluate the superiority of PTD-SQL over existing state-of-the-art frameworks based on the difficulty levels defined by the database, respectively. As depicted in Table~\ref{tab:level}, PTD-SQL outperforms DIN-SQL and DAIL-SQL across different base LLMs, particularly at hard and extra difficulty levels, indicating that LLM can specialize in a problem group and demonstrate enhanced targeted reasoning ability after imitating and delving into problems within the same group.


\begin{figure}[!t]
    \centering
    \hspace{-5mm}
    \subfigure{
        \includegraphics[width=0.48\linewidth]{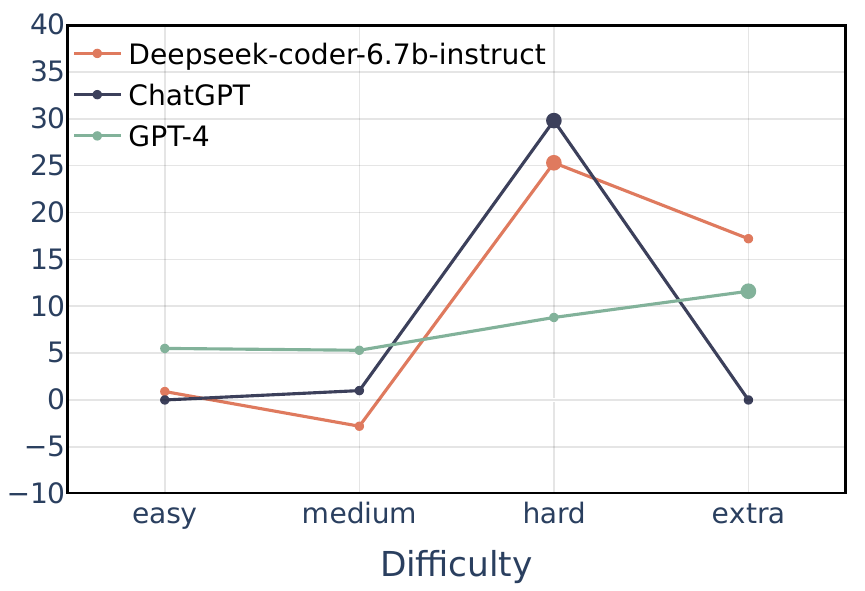}
        \label{fig:without dura}
        }
    \subfigure{
        \includegraphics[width=0.48\linewidth]{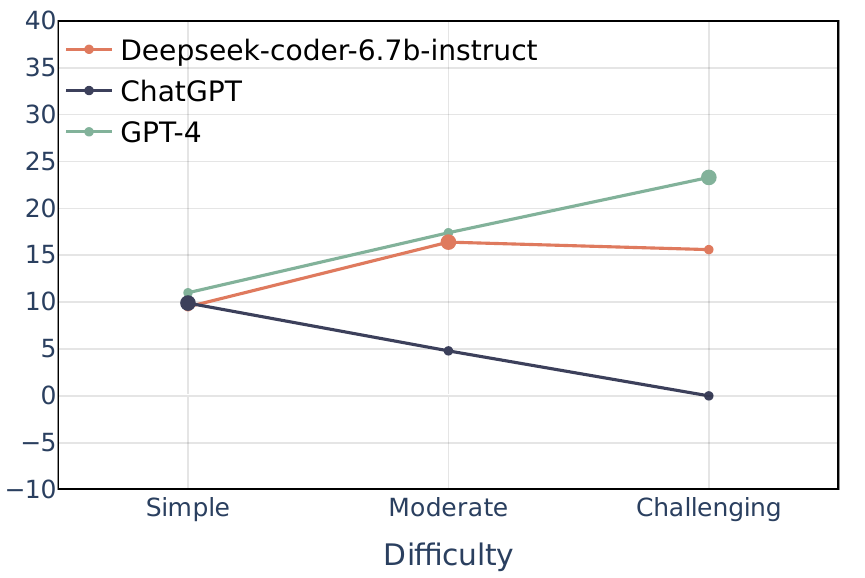}
        \label{fig:valid mrr}
        }
        
        \caption{Under different difficulty levels, the percentage gain~(\%) in EX metric on Spider~(left) and BIRD~(right) obtained by the three models using PTD-SQL compared to DIN-SQL.}
    \label{fig:gain_dif}
\end{figure}

Moreover, we illustrate the performance variations of PTD-SQL in comparison to DIN-SQL across different problem types, thereby discerning the disparities between problem group partitioning strategies and difficulty grading strategies. As inferred from Figure~\ref{fig:gain_dif}, LLMs have made great progress at their respective capacity limits under PTD-SQL. For instance, ChatGPT, akin to a diligent student, achieves a 29.8\% improvement in hard difficulty by focusing on similar problems but fails to progress in extra difficulty, possibly due to inherent model limitations. The deepseek-coder-6.7b-instruct model, with capabilities comparable to ChatGPT, also shows the most significant improvement in hard difficulty (25.3\% vs 17.2\% on extra). However, GPT-4, resembling an elite student, achieves the most substantial breakthrough in extra difficulty and refines its responses across other difficulty levels through referencing and absorption. The results on the BIRD dataset show that GPT-4 achieves the largest increase in performance in the challenging group, while the other two models focus on simple and moderate difficulties. This suggests that LLMs with different levels of reasoning capability can guarantee their upper limit by practicing questions. Detailed results on BIRD are depicted in Appendix~\ref{app:supp_diff}.

\begin{table}[!t]
    \centering
    \resizebox{1.0\linewidth}{!}{
    \begin{tabular}{c|c|c|c|c|c|c}
      \toprule
      Model & QGP Method & Easy & Medium & Hard & Extra & All \\
      \midrule
      \multirow{3}{*}{ChatGPT} & 
      w/o QGP & 84.7 & 76.5 & 71.8 & 52.4& 73.8\\
      & ChatGPT + 10-shot & 86.7 & 78.9 & 74.1 & 56.0 & 76.3\\
      & Llama-2-7b + LoRA & \textbf{90.7} & \textbf{83.1} & \textbf{80.6} & \textbf{56.6} & \textbf{80.3} \\
      \midrule
      \multirow{3}{*}{\shortstack{Deepseek-coder\\-6.7b-instruct}} & 
      w/o QGP & 84.7 & 76.5 & 71.8 & 52.4& 73.8\\
      & ChatGPT + 10-shot & 84.3 & \textbf{79.1} & 69.5 & 54.8 & 74.9\\
      & Llama-2-7b + LoRA & \textbf{87.1} & 78.9 & \textbf{74.9} & \textbf{57.2} & \textbf{76.7} \\
      \bottomrule
    \end{tabular}
    }
    \caption{EX performance based on partition with different accuracy levels on the Spider-dev dataset.}
    \vspace*{-3mm}
    \label{tab:qgp_ablation}
  \end{table}

\begin{figure*}[!t]
    \centering
    \resizebox{0.9\linewidth}{!}{
    \includegraphics[]{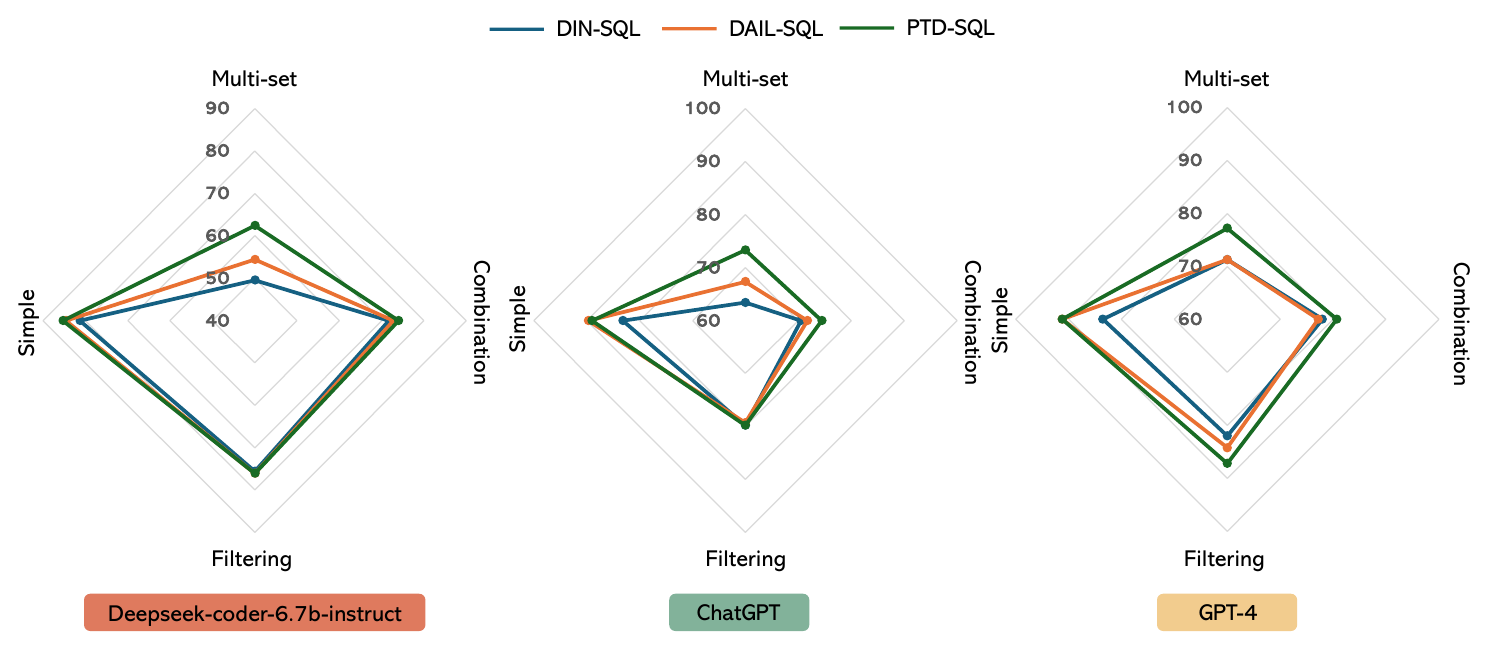}
    }
    \caption{EX of three LLMs on Spider-dev dataset when equipped with DIN-SQL, DAIL-SQL and PTD-SQL.}
    \label{fig:group_ablation}
\end{figure*}

\subsection{RQ2: Performance under Problem Groups}

As depicted in Figure~\ref{fig:group_ablation}, PTD-SQL demonstrates a more pronounced advantage in multi-set problems and combination problems when employing three different baseline models. These problem types entail more intricate reasoning and perplexing conditions. Apart from when using GPT-4, the other two models yield very similar results in the filtering problem across the three methods. This suggests that this category of problem relies more on the inherent ability of the model to effectively organize the filtering conditions rather than emphasizing the logical level. Besides, we consider the detailed performance of queries with multiple question type features in Appendix~\ref{app:sec:fine_grained}, and propose findings and directions for further improvement.

\subsection{RQ3: Effectiveness of QGP}

In this section, we examine the impact of the QGP subtask. As shown in Table~\ref{tab:qgp_result}, the Few-shot method does not align well within a specific context, resulting in weaker performance compared to the fine-tuned model. To further investigate this, we conduct additional experiments involving problem groups classified by ChatGPT, as well as experiments that eliminate the QGP stage and directly recall shots from all targeted drilling banks. The findings presented in Table~\ref{tab:qgp_ablation} indicate that a decline in QGP accuracy adversely affects the final outcomes, with a relative decrease of 5.0\% when testing on ChatGPT. Besides, ChatGPT exhibits a slight reduction in extra difficulty, while Deepseek demonstrates tolerance for classification accuracy at medium to easy difficulty levels. However, upon removing the QGP, the model surpasses the zero-shot performance, but there is a substantial decline in the results. This observation implies that incorporating various types of questions during similarity retrieval might introduce confusion and burden to the model and also validate the relevance of the QGP stage.

\subsection{RQ4: Ablation on Few-shot Selection}
\begin{figure}[!t]
    \centering
    \resizebox{\linewidth}{!}{
    \includegraphics[]{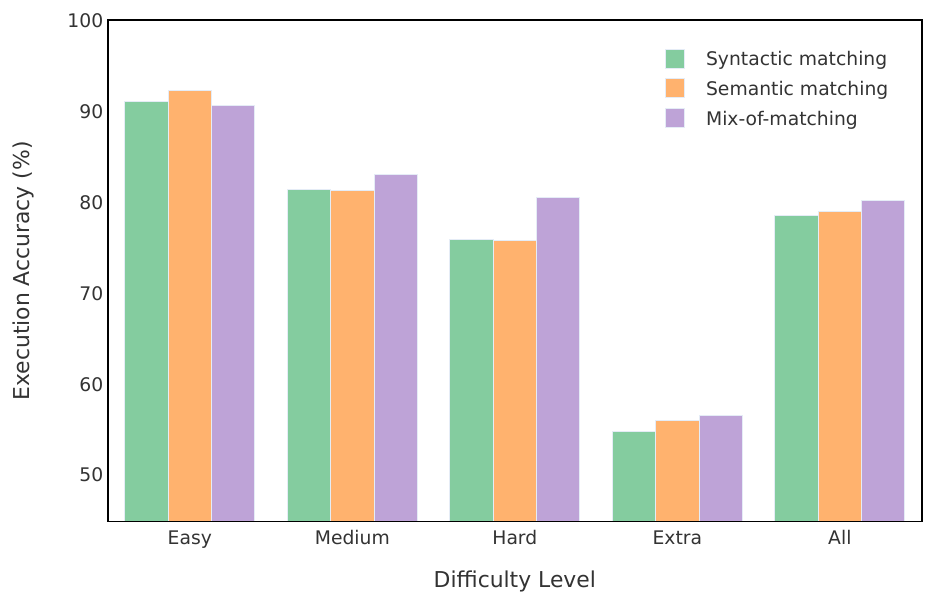}
    }
    \caption{Ablations on few-shot selection strategy on the Spider-dev dataset.~(Detailed data in Table~\ref{tab:supp_selection_ablation})}
    \label{fig:selection_ablation}
    \vspace*{-3mm}
\end{figure}

In this section, ablation experiments are conducted for three distinct shot selection strategies within the same problem group. As illustrated in Figure~\ref{fig:selection_ablation}, the hybrid strategy demonstrates a favorable integration effect beyond the 'easy' category, resulting in an overall improvement. This finding suggests that considering both query keywords and semantic similarity can yield a more comprehensive prompting effect.

\subsection{Ablation on Few-shot Effect}

\begin{table}[b]
    \centering
    \resizebox{1.0\linewidth}{!}{
    \begin{tabular}{c|c|c|c|c|c}
      \toprule
      Few-shot & Easy & Medium & Hard & Extra & All \\
      \midrule
      1-shot & 89.1&76.6&64.9&\underline{56.0}&74.4\\
      2-shot & \underline{89.9}&\underline{80.2}&\underline{72.0}&55.4&\underline{77.2}\\
      4-shot & \textbf{90.7}&\textbf{83.1}&\textbf{80.6}&\textbf{56.6}&\textbf{80.3}\\
      \bottomrule
    \end{tabular}
    }
    \caption{EX on different numbers of few-shot samples w.r.t difficulty-level.}
    \label{tab:shots_diff}
  \end{table}

\begin{table}[ht]
    \centering
    \resizebox{1.0\linewidth}{!}{
    \begin{tabular}{c|c|c|c|c}
      \toprule
      Few-shot & Multi-set & Combination & Filtering & Other Simple Problem\\
      \midrule
      1-shot & 64.4&65.0&74.6&\underline{86.4}\\
      2-shot & \underline{66.3}&\underline{72.8}&\underline{76.6}&86.1\\
      4-shot & \textbf{73.3}&\textbf{74.4}&\textbf{79.8}&\textbf{89.0}\\
      \bottomrule
    \end{tabular}
    }
    \caption{EX on different numbers of few-shot samples w.r.t problem groups.}
    \label{tab:shots_group}
  \end{table}

As a few-shot prompting method, we believe that the number of examples is also an important factor affecting the results. Due to the context limitations we mentioned, we conduct ablation experiments with 4 shots or fewer. For the 1-shot scenario, we selected the single most similar example based on semantic similarity. The performance of PTD-SQL under different numbers of examples, different difficulty levels, and different question types is shown in Table~\ref{tab:shots_diff} and Table~\ref{tab:shots_group}, respectively.

Our results show that when the number of examples is small, it has a greater impact on the final results, and the EX indicator generally shows a growing trend with the increase of examples. This suggests that more examples can stimulate more diverse thinking abilities under relatively limited context constraints. In our framework, more examples mean that the model has done more research on the same type of questions, thus achieving better results.

\section{Conclusion}

In this article, a novel method called PTD-SQL is proposed for LLMs to conduct targeted drilling on specific groups of questions after partitioning. This approach addresses the category tendency of SQL queries, which has been overlooked in previous work. By focusing on the thinking logic of specific types, LLM can effectively enhance its reasoning capabilities. Empirical observations from our comprehensive ablation studies reveal that PTD-SQL significantly reduces the likelihood of LLM making errors within its distinct capability range while demonstrating substantial gains across various question groups. Furthermore, it is posited that this approach can be extended to other domains, such as math word problems and different types of code problems, paving the way for future research.

\section{Limitations}

The limitations of this article lie in the exploration of its effectiveness on larger-scale databases with a broader domain span. Moreover, even SQL statements with strong structural characteristics may have different types of divisions. Therefore, a more detailed investigation of performance under these different divisions can be further improved and optimized. Besides, as stated in Appendix~\ref{app:sec:fine_grained}, for queries with multiple question types, we can also recall example questions from multiple shot banks to comprehensively consider the model and improve the fault tolerance of QGP subtasks. This may be an interesting topic that can be improved in the future. In addition, due to space constraints, this article doesn't optimize for more detailed issues such as schema linking and database content alignment. However, the optimization methods for these issues can be relatively easily integrated into PTD-SQL as a downstream optimization method. Due to our greater focus on the improvement of LLM's reasoning ability for the question answering itself in this article, we are confident that we can achieve better results by adding the aforementioned sub-optimization methods.

\bibliography{custom}

\appendix
\onecolumn
\newpage
\setcounter{tocdepth}{0}
\addtocontents{toc}{\protect\setcounter{tocdepth}{3}}

\renewcommand{\contentsname}{Appendices Content}  
\hypersetup{linkcolor=teal}
\renewcommand{\cftsecleader}{\cftdotfill{1}} 
\renewcommand{\cftsecfont}{\normalfont} 
\renewcommand{\cftsecpagefont}{\normalfont} 
\renewcommand{\cftsecdotsep}{4.5} 
\renewcommand{\cftsubsecdotsep}{4.5} 
\renewcommand{\cftsubsecnumwidth}{2em} 
\renewcommand{\cftsubsecindent}{1.5em} 
\tableofcontents
\hypersetup{linkcolor=blue}
\thispagestyle{empty}
\newpage

\section{Supplementary Statistics}
\subsection{Statistics of Targeted Drilling Banks}
\label{app:sec:bank_statistics}

On Spider-dev and Spider-realistic datasets, the samples from the four different targeted drilling banks all come from random selections within their respective categories after automated classification in the training set, as shown in Table~\ref{tab:bank_statistics}.

However, the BIRD dataset does not provide a training set with regular attributes for generating candidate question banks. Our testing criterion is to randomly divide the Spider-dev dataset into 20\% for training and the remaining 80\% as a testing benchmark at three different difficulty levels. The training set is used for fine-tuning the classifier and building the targeted drilling bank. The statistical data of the targeted drilling bank on the BIRD-dev dataset is shown in Table A. Additionally, due to the lack of clearly defined multi-set operation queries in the BIRD-dev dataset, we only need to investigate the remaining three question types.

\begin{table}[ht]
    \centering
    \begin{tabular}{l|c|c|c|c}
      \toprule
      Bank Group & Multi-set Problem & Combination Problem& Filtering Problem& Other Simple Problem \\
      \midrule
      Number & 200 & 518 & 377 & 500\\
    
      \bottomrule
    \end{tabular}
    \caption{Statistics of targeted drilling banks on Spider-dev and Spider-realistic datasets.}
    \label{tab:bank_statistics}
  \end{table}

\begin{table}[ht]
    \centering
    \begin{tabular}{l|c|c|c}
      \toprule
      Bank Group & Combination Problem& Filtering Problem& Other Simple Problem \\
      \midrule
      Number & 61 & 234 & 11 \\
    
      \bottomrule
    \end{tabular}
    \caption{Statistics of targeted drilling banks on BIRD-dev datasets.}
    \label{tab:bank_bird_statistics}
  \end{table}

\subsection{Statistics of Employed Benchmark}
\label{app:data_statistics}

The two datasets we use, Spider-dev and Spider-realistic, both have native difficulty levels defined by the database itself. The specific data is shown in Table~\ref{tab:data_statistics}.

\begin{table}[ht]
    \centering
    \begin{tabular}{l|c|c|c|c|c}
      \toprule
      Dataset & Easy & Medium & Hard & Extra &All\\
      \midrule
      Spider-dev~\cite{spider} & 248 & 446 & 174 & 166 & 1034\\
      Spider-realistic~\cite{spider_real} & 109 & 203 & 99 & 97 & 508\\
    
      \midrule
      Dataset & Simple & Moderate & Challenging & All & \  \\
      \midrule
      BIRD-dev & 925 & 465 & 144 & 1534 & \ \\
      \bottomrule
    \end{tabular}
    \caption{Statistics of employed three benchmarks.}
    \label{tab:data_statistics}
  \end{table}

\newpage
\section{Supplementary Results}
\label{app:sec:supp_results}

\subsection{Percentage Gain under Different Difficulty-level}\label{app:supp_diff}

The specific data of Figure~\ref{fig:gain_dif}~(left) is shown in Table~\ref{tab:supp_gain_dif}. Correspondingly, data on BIRD-dev~(right) is demonstrated in Table~\ref{tab:supp_bird_gain_dif}. 

We also provide the detailed results on the BIRD-dev dataset split by difficulty in Table~\ref{tab:supp_bird_diff}.

\begin{table}[!h]
    \centering
    \begin{tabular}{c|c|c|c|c|c}
      \toprule
      Base Model & Method &  Simple & Moderate & Challenging & All \\
      \midrule
      \multirow{3}{*}{\shortstack{Deepseek-coder\\-6.7b-instruct}} & DIN-SQL & 47.0 & 32.6 & 27.1 & 40.7  \\
             & DAIL-SQL & 46.3 & 37.7 & 32.2 & 42.4 \\
             & PTD-SQL & 51.5 & 38.0 & 31.4 & 45.4 \\
        \midrule
        \multirow{3}{*}{ChatGPT} & DIN-SQL & 46.6 & 33.4 & 29.7 & 41.0 \\
             & DAIL-SQL & 45.7 & 36.4 & 28.8 & 41.2 \\
             & PTD-SQL & 51.2 & 35.0 & 29.7 & 44.2 \\
        \midrule
        \multirow{3}{*}{GPT-4} & DIN-SQL & 56.9 & 41.4 & 36.4 & 50.2 \\
             & DAIL-SQL & \underline{60.7} & \underline{43.1} & \underline{42.4} & \underline{53.6} \\
             & PTD-SQL & \textbf{63.2} & \textbf{48.7} & \textbf{44.9} & \textbf{57.0} \\
    
      \bottomrule
    \end{tabular}
    \caption{Performance comparison on three LLMs across difficulty levels on BIRD-dev dataset. The best results under each difficulty level when using different LLMs are addressed by \textbf{bold}.}
    \label{tab:supp_bird_diff}
  \end{table}

\begin{table}[ht]
    \centering
    \begin{tabular}{l|c|c|c|c}
      \toprule
      Base Model & Easy & Medium & Hard & Extra \\
      \midrule
      Deepseek-coder-6.7b-instruct & {\color{darkgreen} 0.9} & {\color{peach} 2.8} & {\color{darkgreen} 25.3}& {\color{darkgreen} 17.2}\\
      ChatGPT & {\color{peach}0.0} & {\color{darkgreen}1.0} & {\color{darkgreen}29.8} & {\color{peach}0.0} \\
      GPT-4 & {\color{darkgreen}5.5} & {\color{darkgreen}5.3} & {\color{darkgreen}8.8} & {\color{darkgreen}11.6} \\
    
      \bottomrule
    \end{tabular}
    \caption{EX percentage gain on Spider-dev when compared to DIN-SQL. Number with {\color{darkgreen} green} means an increase, while {\color{peach} red} means a decrease~(\%) or no change.}
    \label{tab:supp_gain_dif}
  \end{table}

\begin{table}[ht]
    \centering
    \begin{tabular}{l|c|c|c}
      \toprule
      Base Model & Simple & Moderate & Challenging \\
      \midrule
      Deepseek-coder-6.7b-instruct & {\color{darkgreen} 9.5} & {\color{darkgreen} 16.4} & {\color{darkgreen} 15.6}\\
      ChatGPT & {\color{darkgreen}9.9} & {\color{darkgreen}4.8} & {\color{peach}0.0} \\
      GPT-4 & {\color{darkgreen}11.0} & {\color{darkgreen}17.4} & {\color{darkgreen}23.3} \\
    
      \bottomrule
    \end{tabular}
    \caption{EX percentage gain on BIRD-dev when compared to DIN-SQL. Number with {\color{darkgreen} green} means an increase, while {\color{peach} red} means a decrease~(\%) or no change.}
    \label{tab:supp_bird_gain_dif}
  \end{table}

\subsection{Performance under Different Problem Groups}

The detailed data of Figure~\ref{fig:group_ablation} is demonstrated in Table~\ref{tab:supp_group_ablation}.

\begin{table}[ht]
    \centering
    \resizebox{0.8\linewidth}{!}{
    \begin{tabular}{c|c|c|c|c|c}
      \toprule
      Base Model & Method & Multi-set & Combination & Filtering & Other Simple Problem \\
      \midrule
      \multirow{3}{*}{\shortstack{Deepseek-coder\\-6.7b-instruct}} & DIN-SQL & 49.5 & 71.7 & 75.6 & 81.3 \\
             & DAIL-SQL & 54.4& 72.8& 76.1& 84.6 \\
             & PTD-SQL & 62.4 & 74.0& 76.1& 85.3 \\
        \midrule
        \multirow{3}{*}{ChatGPT} & DIN-SQL & 63.4&70.5&79.8&83.2 \\
             & DAIL-SQL & 67.3&71.7&79.3&\underline{89.7}\\
             & PTD-SQL & \underline{73.3}&74.4&79.8&89.0 \\
        \midrule
        \multirow{3}{*}{GPT-4} & DIN-SQL & 71.3&\underline{78.0}&82.0&83.5 \\
             & DAIL-SQL & 71.3&77.2&\underline{84.2}&\textbf{91.2} \\
             & PTD-SQL & \textbf{77.2}&\textbf{80.7}&\textbf{87.2}&\textbf{91.2}\\
    
      \bottomrule
    \end{tabular}
    }
    \caption{Detailed EX accuracy of three methods on Spider-dev dataset split by problem groups. }
    \label{tab:supp_group_ablation}
  \end{table}

\subsection{Discussion on Choice of Embeddings}

Many transformer-based models are widely used for text embedding. We select the relatively outstanding all-MiniLM-L6-v2\footnote{\href{https://huggingface.co/sentence-transformers/all-MiniLM-L6-v2}{https://huggingface.co/sentence-transformers/all-MiniLM-L6-v2}} and sentence-t5-large\footnote{\href{https://huggingface.co/sentence-transformers/sentence-t5-large}{https://huggingface.co/sentence-transformers/sentence-t5-large}} for comparison. We ensure that the data usage and process are completely consistent, merely replacing the embedding model used in Section 3.3. The results on Spider are shown in Table~\ref{tab:embedding_choice}. 

The effects presented are quite similar. Upon sample observation, the content retrieved by the model is also similar, which may be related to the dataset. It could also be due to the fact that SQL text queries are essentially composed of purpose statements and database-related items, thus leading to a high degree of differentiation. As for embedding model, text-embedding-ada-002 demonstrates good performance on benchmarks like BEIR, and due to its low cost, it is a very economical choice. 

\begin{table}[ht]
    \centering
    \begin{tabular}{c|c|c}
        \toprule
        Base Model & Embedding Model & EX \\
        \midrule
        \multirow{3}{*}{Deepseek-coder-6.7b-instruct} & openai-text-embedding-ada-002& 75.4 \\
        & all-MiniLM-L6-v2 & 74.5 \\
        & sentence-t5-large & 74.8 \\
        \midrule
        \multirow{3}{*}{ChatGPT} & openai-text-embedding-ada-002& 79.0 \\
        & all-MiniLM-L6-v2 & \textbf{79.3} \\
        & sentence-t5-large & 78.8 \\
        \bottomrule
    \end{tabular}
    \caption{Experiments using Deepseek-coder-6.7b-instruct and ChatGPT with different embedding models.}
    \label{tab:embedding_choice}
\end{table}

\subsection{Detailed Performance of Shots Selection Ablation}

We implement detailed EX performance under different shots auto-selection strategy in Table~\ref{tab:supp_selection_ablation}. We additionally select random selection as a baseline.

\begin{table}[ht]
    \centering
    \begin{tabular}{c|c|c|c|c|c}
      \toprule
      Model & Easy & Medium & Hard & Extra & All \\
      \midrule
      PTD-SQL + Random Selection & 90.3 & 79.8 & 73.4 & 53.6 & 77.1 \\
      PTD-SQL + Syntactic Matching & \underline{91.1} & \underline{81.5} & \underline{76.0} & 54.8 & 78.6 \\
      PTD-SQL + Semantic Matching & \textbf{92.3} & 81.3 & 75.9 & \underline{56.0} & \underline{79.0}\\
      PTD-SQL + Mix-of-Matching & 90.7 & \textbf{83.1} & \textbf{80.6} & \textbf{56.6} & \textbf{80.3} \\
      \bottomrule
    \end{tabular}
    \caption{Ablation study on different few-shot auto-selection strategies on Spider-dev dataset. We employ ChatGPT as reasoning LLM.}
    \label{tab:supp_selection_ablation}
  \end{table}

We also conduct a comparison using GPT-4 which is shown in Table \ref{tab:supp_selection_ablation_gpt4}. However, the difference between random selection and mix-of-matching is minimal. This indicates that on the current benchmark, QGP can significantly alleviate the technical requirements on example selection, as the randomly selected example data already demonstrates considerable relevance.

\begin{table}[ht]
    \centering
    \begin{tabular}{c|c|c|c|c|c}
      \toprule
      Model & Easy & Medium & Hard & Extra & All \\
      \midrule
      PTD-SQL + Random Selection & 94.4	&88.8	&83.3	&62.7	&85.0 \\
      PTD-SQL + Mix-of-Matching &94.8	&88.8&	85.1&	64.5	&85.7 \\
      \bottomrule
    \end{tabular}
    \caption{Ablation study on different few-shot auto-selection strategies on Spider-dev dataset. We employ GPT-4 as reasoning LLM.}
    \label{tab:supp_selection_ablation_gpt4}
  \end{table}

\subsection{Fine-grained analysis of multiple-type queries}
\label{app:sec:fine_grained}

In this section, we explore the potential constraints arising from the fact that certain questions may fall into multiple question groups. We posit that based on the keyword classification method delineated in Section~\ref{sec:qgp} applied to the training set, we can directly apply this to the test set to derive a potential set of question groups. For instance, the ground-truth SQL query 'SELECT Country FROM singer WHERE Age > 40 INTERSECT SELECT Country FROM singer WHERE Age < 30' is categorized as (Multi-set, Filtering).

We define the set of multiple categories to which each query should belong as X and the single group label Y obtained after the fine-tuned Llama-2-7b model completes the QGP. In Table~\ref{tab:multi_type_fine-grained} and Table~\ref{tab:multi_type_fine-grained_4}, we present the EX for all possible partition sets when using ChatGPT and GPT-4, respectively.

Initially, a generally accurate classification can yield relatively satisfactory results. For instance, queries featuring combination problem and filtering problem characteristics exhibit a commendable EX when they are divided into these two subclasses, given a sufficiently large number of samples. Similarly, queries with multi-set and filtering problem features can also attain comparable and favorable EX indicators when they are divided into their respective groups. This suggests that in most instances, a question with multiple types of tendencies can draw insights from a single question bank and make reasonably accurate inferences.

Nonetheless, certain observations also highlight specific limitations of PTD-SQL. For example, in the case of combination-type questions, superior overall results were achieved when they were directly classified as simple problems. This is because these questions, in contrast to those classified as combination problems, contain a greater number of easy and medium-difficulty problems, thus directly benefiting from the simplicity of CoT. Consequently, for future optimization of PTD-SQL, it could be considered, as suggested in the DIN-SQL method, to incorporate the difficulty of the query, thereby preventing some simple questions from being disrupted by complex thought processes.

When comparing the data between Table 1 and Table 2, we can find that GPT-4's stronger fundamental reasoning ability allows for a greater tolerance for misclassification risks. At the same time, the gap in performance between Combination-type questions classified as simple questions and those classified as Combination itself is also reduced.

\begin{table*}[!h]
    \centering
    \resizebox{1.0\linewidth}{!}{
    \begin{tabular}{l|cc|cc|cc|cc}
      \toprule
      \multirow{2}{*}{} & \multicolumn{2}{|c}{Combination Problem} & \multicolumn{2}{|c}{Multi-set Problem} & \multicolumn{2}{|c}{Filtering Problem} & \multicolumn{2}{|c}{Simple Problem} \\
      & Num & EX & Num & EX & Num & EX & Num & EX \\
      \midrule
      (Combination,) & 236 & 60.6& - & - & 18 & 44.4& 100 & 86.0\\
      (Combination, Filtering,) & 12 & 83.3 & 2 & 0.0 & 27 & 44.4 & 6 & 50.0\\
      (Combination, Multi-set,) & - & - & 2 & 0.0 & - & - & - & - \\
      (Combination, Multi-set, Filtering,) & - & -& 4 & 50.0 & - & -& -& -\\
      (Multi-set,) & - & - & 18 & 61.1 & 2 & 50.0 & - & -\\
      (Multi-set, Filtering,) & - & - & 50 & 72.0 & 4 & 100.0 & - & - \\
      (Filtering,) & 3 & 100.0 & 25 & 52.0 & 349 & 79.4 & 12 & 83.3 \\
      (Simple,) & 3 & 33.3 & - & - & 6 & 33.3 & 155 & 87.7 \\ 
      \bottomrule
    \end{tabular}
    }
    \caption{Fine-grained EX results of ambiguity in question types when using ChatGPT.}
    \label{tab:multi_type_fine-grained}
  \end{table*}

\begin{table*}[!h]
    \centering
    \resizebox{1.0\linewidth}{!}{
    \begin{tabular}{l|cc|cc|cc|cc}
      \toprule
      \multirow{2}{*}{} & \multicolumn{2}{|c}{Combination Problem} & \multicolumn{2}{|c}{Multi-set Problem} & \multicolumn{2}{|c}{Filtering Problem} & \multicolumn{2}{|c}{Simple Problem} \\
      & Num & EX & Num & EX & Num & EX & Num & EX \\
      \midrule
      (Combination,) & 236 & 79.7 & - & - & 18 & 66.7 & 100 & 91.0\\
      (Combination, Filtering,) & 12 & 75.0 & 2 & 50.0 & 27 & 66.7 & 6 & 100.0\\
      (Combination, Multi-set,) & - & - & 2 & 50.0 & - & - & - & - \\
      (Combination, Multi-set,Filtering,) & - & -& 4 & 50.0 & - & -& -& -\\
      (Multi-set,) & - & - & 18 & 72.2 & 2 & 100.0 & - & -\\
      (Multi-set, Filtering,) & - & - & 50 & 78.0 & 4 & 75.0 & - & - \\
      (Filtering,) & 3 & 100.0 & 25 & 60.0 & 349 & 90.0 & 12 & 83.3 \\
      (Simple,) & 3 & 100.0 & - & - & 6 & 83.3 & 155 & 91.0 \\ 
      \bottomrule
    \end{tabular}
    }
    \caption{Fine-grained EX results of ambiguity in question types when using GPT-4.}
    \label{tab:multi_type_fine-grained_4}
  \end{table*}

\newpage
\subsection{Case Study}
\label{app:sec:case_study}

In this section, we give some detailed cases to show how PTD-SQL assists LLMs in Text-to-SQL. As shown in Table~\ref{tab:complex_case}, the output of DIN-SQL shows that DIN-SQL doesn't properly handle multi-set queries. Simply using 'OR' for connection may cause conflict issues in multi-table queries. However, PTD-SQL successfully solves the problem by properly learning the decomposition of the problem and using corresponding keywords to connect subqueries correctly. In addition, as shown in Table~\ref{tab:combination_case}, the sentences generated by DIN-SQL may have the risk of confusing filtering phrases, while PTD-SQL teaches LLM to think hierarchically about filtering logic and answer organization, thus leading to the correct results.

\begin{table*}[htbp]
	\centering
	\begin{tabularx}{\textwidth}{lX}
		\toprule
	   \textbf{Query} & Find all airlines that have flights from both airports 'APG' and 'CVO'.  \\
		\midrule
		\textbf{DIN-SQL output} 
		& SELECT Airline  FROM airlines  JOIN flights ON airlines.uid = flights.Airline  WHERE flights.SourceAirport = 'APG' OR flights.SourceAirport = 'CVO'
		\\
		\midrule
		\textbf{PTD-SQL output} & SELECT Airline FROM airlines WHERE EXISTS (SELECT * FROM flights JOIN airports ON flights.SourceAirport = airports.AirportCode WHERE airports.AirportCode = 'APG' AND airlines.uid = flights.Airline) INTERSECT SELECT Airline FROM airlines WHERE EXISTS (SELECT * FROM flights JOIN airports ON flights.SourceAirport = airports.AirportCode WHERE airports.AirportCode = 'CVO' AND airlines.uid = flights.Airline) \\
        \midrule
        \textbf{Reasoning} & <1> Question Decomposition: In this step, we contemplate how to decompose the query. The query emphasizes the intersection set logic, so we can decompose the question into two subproblems: 1. find all airlines that have flights from airport 'APG'; 2. find all airlines that have flights from airport 'CVO'. <2> Schema Linking: In this step, we identify the tables and columns that should be used based on the requirements of the query and the foreign key relationships. To complete both subproblems, we need to use tables ['airlines', 'flights', 'airports']. <3> Operation: Use 'where' to filter using column 'SourceAirport' in table 'flights' and column 'AirportCode' in table 'airports'. <4> SQL Generation: Use the 'intersect' operation to connect the queries of subproblems to form the final SQL statement. \\
		\bottomrule
	\end{tabularx}
	\caption{A detailed case of a complex problem on the Spider-dev dataset when using ChatGPT.}
 \label{tab:complex_case}
\end{table*}

\begin{table*}[htbp]
	\centering
	\begin{tabularx}{\textwidth}{lX}
		\toprule
	   \textbf{Query} & Which countries in Europe have at least 3 car manufacturers?  \\
		\midrule
		\textbf{DIN-SQL output} 
		& SELECT CountryName  FROM countries  WHERE Continent = 'Europe'  AND CountryId IN (SELECT Country FROM car\_makers GROUP BY Country HAVING COUNT(*) >= 3)
		\\
		\midrule
		\textbf{PTD-SQL output} & SELECT CountryName FROM car\_makers  JOIN countries ON car\_makers.Country = countries.CountryId  JOIN continents ON countries.Continent = continents.ContId  GROUP BY Country  HAVING COUNT(*) >= 3  AND continents.Continent = 'Europe' \\
        \midrule
        \textbf{Reasoning} & <1> Operation: The query requires the countries in Europe that have at least 3 car manufacturers. So we should apply the 'count' operation to the 'Id' column in the 'car\_makers' table and find the countries with a count of at least 3. Since the unit to which the car manufacturers are being counted in the query belongs in the country, we should apply the 'group by' operation to the 'Country' column in the 'car\_makers' table.<2> Schema Linking: In this step, we identify the tables and columns that should be used based on the first step and the foreign key relationships. In this question, we need to use tables ['car\_makers', 'countries', 'continents'].<3> SQL Generation: The query requires the countries in Europe that have at least 3 car manufacturers, so we should select the 'CountryName' column from the 'countries' table. We also need to join the 'car\_makers' table with the 'countries' table and the 'continents' table to ensure that we are only considering countries in Europe.\\
		\bottomrule
	\end{tabularx}
	\caption{A detailed case of combination problem on Spider-dev dataset when using ChatGPT.}
 \label{tab:combination_case}
\end{table*}

\section{Use of Evaluation Program}

Following previous work, we employ a widely-used program to attain EX, which is released by~\cite{eval}. The script used on Spider-dev and Spider-realistic datasets is "python3 evaluation.py --gold gold\_path  --pred prediction\_path  --db spider/database/ --table spider/tables.json --etype all --plug\_value --keep\_distinct".

\section{Time and Token cost}
\label{app:sec:time_token_cost}

In this section, we highlight the comparative benefits of PTD-SQL over alternative frameworks concerning time and token usage. Owing to our approach necessitating only a single query, we gain a considerable edge in token efficiency while simultaneously ensuring effective time management and exceptional outcomes. Although certain optimizations aimed at addressing difficulty granularity and schema linking could potentially enhance PTD-SQL's performance, they would unavoidably result in increased time and token expenditures. The detailed comparison is demonstrated in Table.~\ref{tab:time_token_cost}. Data of previous methods are from~\cite{xie2024decomposition}.

\begin{table}[ht]
    \centering
    \begin{tabular}{c|c|c|c}
      \toprule
      Method & Tokens per Query & Inference Time per Query & EX \\
      \midrule
      C3 & 2803 & 19.34s & \textbf{81.2} \\
      DIN-SQL & 9126 & \underline{4.37s} & 76.8 \\
      DAIL-SQL & \textbf{700} & - & 79.1 \\
      \midrule
      PTD-SQL & \underline{1855} & \textbf{3.34s} & \underline{80.3} \\
      \bottomrule
    \end{tabular}
    \caption{Tokens and time cost comparison using ChatGPT.}
    \label{tab:time_token_cost}
  \end{table}

\newpage

\section{Prompt Design}
\label{app:prompt}
In this section, we elaborate on the prompt design employed in our study, which is crucial for the effective application of Large Language Models (LLMs) in Text-to-SQL tasks. The prompts serve as guiding questions or statements that help the LLMs focus on specific aspects of the problem and facilitate their learning process.

\subsection{Targeted Drilling Bank Auto-construction on Spider}
\label{sec:app_bank_prompt}
This section can be seen as supplementary materials for section~\ref{sec:auto-construction}. We provide all four types of shots generation prompts on the Spider dataset, which are leveraged on Spirder-dev and Spider-realistic datasets.

\begin{xltabular}{1.0\linewidth}{X}
    \hline 
    \multicolumn{1}{c}{\textbf{Shots Generation Prompt of Multi-set Problem}} \\
    \hline
    You are a powerful text-to-SQL reasoner. Currently, I am seeking to transform intricate text queries into analytical statements that simplify the creation of SQL statements, leading to the generation of the final SQL query. Our current focus lies in the category of {\color{peach} multi-set operations}. Please learn from the provided examples, design a detailed plan for the text query, and present the resulting SQL query. \\
    \\
    Example 1:\\
    \#\# Tables:\\
    Table aircraft, columns = [*,aid,name,distance]\\
    Table certificate, columns = [*,eid,aid]\\
    Table employee, columns = [*,eid,name,salary]\\
    Table flight, columns = [*,flno,origin,destination,distance,departure\_date,arrival\_date,price,aid]\\
    \#\# Foreign\_keys:\\
    {[flight.aid = aircraft.aid,certificate.aid = aircraft.aid,certificate.eid = employee.eid]} \\
    \#\# Query:\\
    Show names for all employees who have certificates on both Boeing 737-800 and Airbus A340-300.\\
    Let's think step by step.\\
    <1> Question Decomposition: In this step, we contemplate how to decompose the query. The query emphasizes intersection logic, so we can decompose the question into two subproblems: 1. what are the names of employees who have certificates on Boeing 737-800; 2. what are the names of employees who have certificates on Airbus A340-300.\\
    <2> Schema Linking: In this step, we identify the tables and columns that should be used based on the requirements of the query and the foreign key relationships. To complete the first subproblem, we need to use tables 'employee' and 'aircraft'. since table 'employee' and table 'aircraft' do not have a direct foreign key connection, we need to use tables ['employee', 'certificate', 'aircraft']. To complete the second subproblem, we need to use tables ['employee', 'certificate', 'aircraft'] for the same reason.\\
    <3> Operation: Use 'where' to filter using column 'name' in table 'aircraft'.\\
    <4> SQL Generation: Use the 'intersect' operation to connect the queries of subproblems to form the final SQL statement.\\
    SQL query: SELECT T1.name FROM Employee AS T1 JOIN Certificate AS T2 ON T1.eid  =  T2.eid JOIN Aircraft AS T3 ON T3.aid  =  T2.aid WHERE T3.name  =  "Boeing 737-800" INTERSECT SELECT T1.name FROM Employee AS T1 JOIN Certificate AS T2 ON T1.eid  =  T2.eid JOIN Aircraft AS T3 ON T3.aid  =  T2.aid WHERE T3.name  =  "Airbus A340-300"\\
    \\
    Example 2:\\
    \#\# Tables:\\
    Table station, columns = [*,id,name,lat,long,dock\_count,city,installation\_date]\\
    Table status, columns = [*,station\_id,bikes\_available,docks\_available,time] \\
    Table trip, columns = [*,id,duration,start\_date,start\_station\_name,start\_station\_id,end\_date,\\end\_station\_name,end\_station\_id,bike\_id,subscription\_type,zip\_code] \\
    Table weather, columns = [*,date,max\_temperature\_f,mean\_temperature\_f,min\_temperature\_f,\\max\_dew\_point\_f,mean\_dew\_point\_f,min\_dew\_point\_f,max\_humidity,mean\_humidity,min\_humidity,\\max\_sea\_level\_pressure\_inches,mean\_sea\_level\_pressure\_inches,min\_sea\_level\_pressure\_inches,\\max\_visibility\_miles,mean\_visibility\_miles,min\_visibility\_miles,max\_wind\_Speed\_mph,\\mean\_wind\_speed\_mph,max\_gust\_speed\_mph,precipitation\_inches,cloud\_cover,events,wind\_dir\_degrees,\\zip\_code]\\
    \#\# Foreign\_keys:\\
    {[status.station\_id = station.id]}\\
    \#\# Query:\\
    What are the names of stations that have average bike availability above 10 and are not located in San Jose city?\\
    Let's think step by step.\\
    <1> Question Decomposition: In this step, we contemplate how to decompose the query. The query emphasizes difference set logic, so we can decompose the question into two subproblems: 1. what are the names of stations that have average bike availability above 10; 2. what are the names of stations that are located in San Jose city. \\
    <2> Schema Linking: In this step, we identify the tables and columns that should be used based on the requirements of the query and the foreign key relationships. To complete the first subproblem, we need to use tables ['station', 'status']. To complete the second subproblem, we need to use table ['station'].\\
    <3> Operation: Due to the need for calculating the average bike availability for different stations, we need to perform a 'GROUP BY' operation on the column 'station\_id', filter by performing 'HAVING AVG()' on the column 'bikes\_available'. \\
    <4> SQL Generation: Use the 'except' operation to connect the queries of subproblems to form the final SQL statement.\\
    SQL query: SELECT T1.name FROM station AS T1 JOIN status AS T2 ON T1.id  =  T2.station\_id GROUP BY T2.station\_id HAVING avg(bikes\_available)  >  10 EXCEPT SELECT name FROM station WHERE city  =  "San Jose"\\
    \\
    Example 3:\\
    \#\# Tables:\\
    Table aircraft, columns = [*,aid,name,distance]\\
    Table certificate, columns = [*,eid,aid]\\
    Table employee, columns = [*,eid,name,salary]\\
    Table flight, columns = [*,flno,origin,destination,distance,departure\_date,arrival\_date,price,aid]\\
    \#\# Foreign\_keys:\\
    {[flight.aid = aircraft.aid,certificate.aid = aircraft.aid,certificate.eid = employee.eid]}\\
    \#\# Query:\\
    Show ids for all employees who don't have a certificate.\\
    Let's think step by step.\\
    <1> Question Decomposition: In this step, we contemplate how to decompose the query. The query emphasizes difference set logic, so we can decompose the question into two subproblems: 1. what are the ids of employees who have certificates; 2. what are the ids of all employees.\\
    <2> Schema Linking: In this step, we identify the tables and columns that should be used based on the requirements of the query and the foreign key relationships. To complete the first subproblem, we need to use tables ['employee', 'certificate']. To complete the second subproblem, we only need to use the table 'employee'.\\
    <3> Operation: Use 'where' to filter out the employees who have certificates.\\
    <4> SQL Generation: Use the 'except' operation to connect the queries of subproblems to form the final SQL statement.\\
    SQL query: SELECT eid FROM employee EXCEPT SELECT eid FROM certificate\\
    \\
    Example 4:\\
    \#\# Tables:\\
    Table Campuses, columns = [*, Id, Campus, Location, County, Year]\\
    Table csu\_fees, columns = [*,Campus,Year,CampusFee]\\
    Table degrees, columns = [*,Year,Campus,Degrees]\\
    Table discipline\_enrollments, columns = [*, Campus,Discipline,Year, Undergraduate,Graduate]\\
    Table enrollments, columns = [*,Campus,Year,TotalEnrollment\_AY,FTE\_AY]\\
    Table faculty, columns = [*,Campus,Year,Faculty]\\
    \#\# Foreign\_keys:\\
    {[csu\_fees.Campus = Campuses.Id, degrees.Campus = Campuses.Id,discipline\_enrollments.Campus = Campuses.Id, enrollments.Campus = Campuses.Id, faculty.Campus = Campuses.Id]}\\
    \#\# Query:\\
    Find the name of the campuses that is in Northridge, Los Angeles or in San Francisco, San Francisco.\\
    Let's think step by step.\\
    <1> Question Decomposition: In this step, we contemplate how to decompose the query. The query emphasizes union logic, so we can decompose the question into two subproblems: 1. what are the names of the campuses that are in Northridge, Los Angeles; 2. what are the names of the campuses that are in San Francisco, San Francisco.\\
    <2> Schema Linking: In this step, we identify the tables and columns that should be used based on the requirements of the query and the foreign key relationships. To complete both subproblems, we need to use table 'Campuses'.\\
    <3> Operation: Use 'where' to filter using column 'Location' in table 'Campuses'.\\
    <4> SQL Generation: Use the 'union' operation to connect the queries of subproblems to form the final SQL statement.\\
    SQL query: SELECT Campus FROM Campuses WHERE Location  =  "Northridge, Los Angeles" UNION SELECT Campus FROM Campuses WHERE Location  =  "San Francisco, San Francisco"\\
    \hline
    \caption{The prompt used for generating targeted drilling bank shots under \textbf{multi-set problem} group.}\label{tab:multiset_prompt}
    \end{xltabular}

\begin{xltabular}{1.0\linewidth}{X}
    \hline 
    \multicolumn{1}{c}{\textbf{Shots Generation Prompt of Combination Problem}} \\
    \hline
    You are a powerful text-to-SQL reasoner. Currently, I am seeking to transform intricate text queries into analytical statements that simplify the creation of SQL statements, leading to the generation of the final SQL query. Our current focus lies in the category of {\color{peach} combination operations}. Please learn from the provided examples, design a detailed plan for the text query, and present the resulting SQL query. \\
    \\
    Example 1:\\
    \#\# Tables:\\
    Table gymnast, columns = [*,Gymnast\_ID,Floor\_Exercise\_Points,Pommel\_Horse\_Points,Rings\_Points,\\
    Vault\_Points,Parallel\_Bars\_Points,Horizontal\_Bar\_Points,Total\_Points]\\
    Table people, columns = [*,People\_ID,Name,Age,Height,Hometown]\\
    \#\# Foreign\_keys:\\
    {[gymnast.Gymnast\_ID = people.People\_ID]}\\
    \#\# Query:\\
    How many gymnasts are from each hometown?\\
    Let's think step by step.\\
    <1> Operation: the query requires the number of gymnasts from each hometown, so we should apply the 'count' operation to table 'gymnast', and it does not need sort. Since the unit to which the gymnasts being counted in the query belong is hometown and only table 'people' has column 'Hometown', so we should apply the 'group by' operation to column 'Hometown' in table 'people'. \\
    <2> Schema Linking: In this step, we identify the tables and columns that should be used based on the first step and the foreign key relationships. Due to the direct foreign key connection between table 'gymnast' and 'people', we need to use tables ['gymnast', 'people']. \\
    <3> SQL Generation: The query requires the number of gymnasts from each hometown, so we should select the count and 'hometown' in people.\\
    SQL query: SELECT T2. Hometown,  COUNT(*) FROM gymnast AS T1 JOIN people AS T2 ON T1.Gymnast\_ID  =  T2.People\_ID GROUP BY T2.Hometown\\
    \\
    Example 2:\\
    \#\# Tables:\\
    Table county, columns = [*,County\_Id,County\_name,Population,Zip\_code]\\
    Table selection, columns = [*,Election\_ID,Counties\_Represented,District,Delegate,Party,First\_Elected,\\
    Committee]\\
    Table party, columns = [*,Party\_ID,Year,Party,Governor,Lieutenant\_Governor,Comptroller,\\
    Attorney\_General,US\_Senate]\\
    \#\# Foreign\_keys:\\
    {[election.District = county.County\_Id,election.Party = party.Party\_ID]}\\
    \#\# Query:\\
    Show the name of each party and the corresponding number of delegates from that party.\\
    Let's think step by step.\\
    <1> Operation: The query requires the name of each party and the corresponding number of delegates from that party, so we should apply the 'count' operation to table 'election' for the 'Delegate' column, and it does not need sorting. Since the unit to which the delegates being counted in the query belong is the party and only table 'party' has the column 'Party', so we should apply the 'group by' operation to column 'Party' in table 'party'.\\
    <2> Schema Linking: In this step, we identify the tables and columns that should be used based on the first step and the foreign key relationships. Due to the direct foreign key connection between table 'election' and 'party'. We need to use tables ['election', 'party'].\\
    <3> SQL Generation: The query requires the name of each party and the corresponding number of delegates from that party, so we should select the 'Party' column in the 'party' table and count the 'Delegate' column in the 'election' table.\\
    SQL query: SELECT T1.Party, COUNT(*) FROM party AS T1 JOIN election AS T2 ON T1.Party\_ID = T2.Party GROUP BY T1.Party\\
    \\
    Example 3:\\
    \#\# Tables:\\
    Table city, columns = [*,City\_ID,Official\_Name,Status,Area\_km\_2,Population,Census\_Ranking]\\
    Table competition\_record, columns = [*,Competition\_ID,Farm\_ID,Rank]\\
    Table farm, columns = [*,Farm\_ID,Year,Total\_Horses,Working\_Horses,Total\_Cattle,Oxen,\\
    Bulls,Cows,Pigs,Sheep\_and\_Goats]\\
    Table farm\_competition, columns = [*,Competition\_ID,Year,Theme,Host\_city\_ID,Hosts]\\
    \#\# Foreign\_keys:\\
    {[farm\_competition.Host\_city\_ID = city.City\_ID,competition\_record.Farm\_ID}\\
    {= farm.Farm\_ID,competition\_record.Competition\_ID = farm\_competition.Competition\_ID]}\\
    \#\# Query:\\
    Show the status of the city that has hosted the greatest number of competitions.\\
    Let's think step by step.\\
    <1> Operation: The query requires the name of each party and the corresponding number of delegates from that party, so we should apply the 'count' operation to table 'election' for the 'Delegate' column, and it does not need sorting. Since the unit to which the delegates being counted in the query belong is the party and only table 'party' has the column 'Party', so we should apply the 'group by' operation to column 'Party' in table 'party'.\\
    <2> Schema Linking: In this step, we identify the tables and columns that should be used based on the first step and the foreign key relationships. Due to the direct foreign key connection between table 'election' and 'party'. We need to use tables ['election', 'party'].\\
    <3> SQL Generation: The query requires the name of each party and the corresponding number of delegates from that party, so we should select the 'Party' column in the 'party' table and count the 'Delegate' column in the 'election' table.\\
    SQL query: SELECT T1.Party , COUNT(*) FROM party AS T1 JOIN election AS T2 ON T1.Party\_ID = T2.Party GROUP BY T1.Party\\
    \\
    Example 4:\\
    \#\# Tables:\\
    Table city, columns = [*,City\_ID,Official\_Name,Status,Area\_km\_2,Population,Census\_Ranking]\\
    Table competition\_record, columns = [*,Competition\_ID,Farm\_ID,Rank]\\
    Table farm, columns = [*,Farm\_ID,Year,Total\_Horses,Working\_Horses,Total\_Cattle,Oxen,Bulls,Cows,\\
    Pigs,Sheep\_and\_Goats]\\
    Table farm\_competition, columns = [*,Competition\_ID,Year,Theme,Host\_city\_ID,Hosts]\\
    \#\# Foreign\_keys:\\
    {[farm\_competition.Host\_city\_ID = city.City\_ID,competition\_record.Farm\_ID= farm.Farm\_ID,}\\
    {competition\_record.Competition\_ID = farm\_competition.Competition\_ID]}\\
    \#\# Query:\\
    Please show the different statuses, ordered by the number of cities that have each.\\
    Let's think step by step.\\
    <1> Operation: The query requires the different statuses ordered by the number of cities that have each status, so we should apply the 'count' operation to the 'city' table for the 'Status' column, and sort it in ascending order. Since the unit to which the statuses being counted in the query belong is the city, we should apply the 'group by' operation to the 'Status' column in the 'city' table.\\
    <2> Schema Linking: In this step, we identify the tables and columns that should be used based on the first step and the foreign key relationships. In this question, we only need to use table ['city'].\\
    <3> SQL Generation: The query requires the different statuses ordered by the number of cities that have each status, so we should select the 'Status' column in the 'city' table. The query does not require the count of cities so it is only used for filtering and not selected. \\
    SQL query: SELECT Status FROM city GROUP BY Status ORDER BY COUNT(*) ASC\\
    \hline
    \caption{The prompt used for generating targeted drilling bank shots under \textbf{combination problem} group.}\label{tab:combination_prompt}
    \end{xltabular}

\begin{xltabular}{1.0\linewidth}{X}
    \hline 
    \multicolumn{1}{c}{\textbf{Shots Generation Prompt of Filtering Problem}} \\
    \hline
    You are a powerful text-to-SQL reasoner. Currently, I am seeking to transform intricate text queries into analytical statements that simplify the creation of SQL statements, leading to the generation of the final SQL query. Our current focus lies in the category of {\color{peach}filtering problem}. Please learn from the provided examples, design a detailed plan for the text query, and present the resulting SQL query.\\
    \\
    Example 1:\\
    \#\# Tables:\\
    Table city, columns = [*,City\_ID,Official\_Name,Status,Area\_km\_2,Population,Census\_Ranking]\\
    Table competition\_record, columns = [*,Competition\_ID,Farm\_ID,Rank]\\
    Table farm, columns = [*,Farm\_ID,Year,Total\_Horses,Working\_Horses,Total\_Cattle,Oxen,Bulls,Cows,\\
    Pigs,Sheep\_and\_Goats]\\
    Table farm\_competition, columns = [*,Competition\_ID,Year,Theme,Host\_city\_ID,Hosts]\\
    \#\# Foreign\_keys:\\
    {[farm\_competition.Host\_city\_ID = city.City\_ID,competition\_record.Farm\_ID = farm.Farm\_ID,}\\
    {competition\_record.Competition\_ID = farm\_competition.Competition\_ID]}\\
    \#\# Query:\\
    Return the hosts of competitions for which the theme is not Aliens?\\
    Let's think step by step.\\
    <1> Decomposition: The query requires filtering on column 'theme', so we should apply the 'where' to column 'theme' and then return the hosts of selected competition. \\
    <2> Schema Linking: In this step, we identify the tables and columns that should be used based on the first step and the foreign key relationships. Since table 'farm\_competition' has columns 'Theme' and 'Hosts', we only need table 'farm\_competition'.\\
    <3> SQL Generation: Directly write the sql using 'where'.\\
    SQL query: SELECT Hosts FROM farm\_competition WHERE Theme != 'Aliens'\\
    \\
    Example 2:\\
    \#\# Tables:\\
    Table Allergy\_Type, columns = [*,Allergy,AllergyType]\\
    Table Has\_Allergy, columns = [*,StuID,Allergy]\\
    Table Student, columns = [*,StuID,LName,Fname,Age,Sex,Major,Advisor,city\_code]\\
    \#\# Foreign\_keys:\\
    {[Has\_Allergy.Allergy = Allergy\_Type.Allergy,Has\_Allergy.StuID = Student.StuID]}\\
    \#\# Query:\\
    How many female students have milk or egg allergies?\\
    Let's think step by step.\\
    <1> Decomposition: Firstly, we filter candidates using column 'Sex' in table 'Student' and column 'Allergy' in table 'Has\_Allergy'. Secondly, we use 'count' to calculate the number of selected female students.\\
    <2> Schema Linking: In this step, we identify the tables and columns that should be used based on the first step and the foreign key relationships. Since table 'Student' and table 'Has\_Allergy' have direct foreign keys, so we need tables ['Student', 'Has\_Allergy'].\\
    <3> SQL Generation: We need to join the 'Student' and 'Has\_Allergy' tables on the 'StuID' column. Then, we filter the rows where 'Sex' is 'F' and 'Allergy' is either 'Milk' or 'Eggs'. Finally, we count the number of rows that meet these conditions.\\
    SQL query: SELECT count(*) FROM has\_allergy AS T1 JOIN Student AS T2 ON T1.StuID  =  T2.StuID WHERE T2.sex = 'F' AND T1.allergy = 'Milk' or T1.allergy = 'Eggs'\\
    \\
    Example 3:\\
    \#\# Tables:\\
    Table station, columns = [*,id,name,lat,long,dock\_count,city,installation\_date]\\
    Table status, columns = [*,station\_id,bikes\_available,docks\_available,time]\\
    Table trip, columns = [*,id,duration,start\_date,start\_station\_name,start\_station\_id,end\_date,\\
    end\_station\_name,end\_station\_id,bike\_id,subscription\_type,zip\_code]\\
    Table weather, columns = [*,date,max\_temperature\_f,mean\_temperature\_f,min\_temperature\_f,\\
    max\_dew\_point\_f,mean\_dew\_point\_f,min\_dew\_point\_f,max\_humidity,mean\_humidity,min\_humidity,\\
    max\_sea\_level\_pressure\_inches,mean\_sea\_level\_pressure\_inches,min\_sea\_level\_pressure\_inches,\\
    max\_visibility\_miles,mean\_visibility\_miles,min\_visibility\_miles,max\_wind\_Speed\_mph,\\
    mean\_wind\_speed\_mph,max\_gust\_speed\_mph,precipitation\_inches,cloud\_cover,events,\\
    wind\_dir\_degrees,zip\_code] \\
    \#\# Foreign\_keys:\\
    {[status.station\_id = station.id]}\\
    \#\# Query:\\
    How many trips did not end in San Francisco?\\
    Let's think step by step.\\
    <1> Decomposition: The query requires filtering on trips that did not end in San Francisco. Firstly, we need to identify the stations located in San Francisco. Secondly, we need to filter trips based on their end\_station\_id.\\
    <2> Schema Linking: In this step, we identify the tables and columns that should be used based on the first step and the foreign key relationships. In the first step, we need to select id from table 'station' where city = 'San Francisco'. In the second step, we need to select id from table 'trip' and filter by end\_station\_id.\\
    <3> SQL Generation: Use 'where' to filter stations in San Francisco, and then use 'not in' to filter trips that did not end in San Francisco.\\
    SQL query: SELECT COUNT(*) FROM trip WHERE end\_station\_id NOT IN (SELECT id FROM station WHERE city = 'San Francisco')\\
    \\
    Example 4:\\
    \#\# Tables:\\
    Table concert, columns = [*,concert\_ID,concert\_Name,Theme,Stadium\_ID,Year]\\
    Table singer, columns = [*,Singer\_ID,Name,Country,Song\_Name,Song\_release\_year,Age,Is\_male]\\
    Table singer\_in\_concert, columns = [*,concert\_ID,Singer\_ID]\\
    Table stadium, columns = [*,Stadium\_ID,Location,Name,Capacity,Highest,Lowest,Average]\\
    \#\# Foreign\_keys:\\
    {[concert.Stadium\_ID = stadium.Stadium\_ID,singer\_in\_concert.Singer\_ID = singer.Singer\_ID,}\\
    {singer\_in\_concert.concert\_ID = concert.concert\_ID]}\\
    \#\# Query:\\
    Find the number of concerts that happened in the stadium with the highest capacity .\\
    Let's think step by step.\\
    <1> Decomposition: Firstly, we need to find the stadium with the highest capacity. Secondly, we need to filter concerts based on their stadium and count them.\\
    <2> Schema Linking: In this step, we identify the tables and columns that should be used based on the first step and the foreign key relationships. In the first step, we need to select stadium\_id with the highest capacity from table 'stadium'. In the second step, we need to filter stadium\_id from table 'concert' and count them.\\
    <3> SQL Generation: Use 'order by' and 'desc' to select stadium with highest capacity, and then use 'where' to filter concert and count it.\\
    SQL query: select count(*) from concert where stadium\_id = (select stadium\_id from stadium order by capacity desc limit 1)\\
    \hline
    \caption{The prompt used for generating targeted drilling bank shots under \textbf{filtering problem} group.}\label{tab:filter_prompt}
    \end{xltabular}

\begin{xltabular}{1.0\linewidth}{X}
    \hline 
    \multicolumn{1}{c}{\textbf{Shots Generation Prompt of Other Simple Problem}} \\
    \hline
    You are a powerful text-to-SQL reasoner. Currently, I am seeking to transform intricate text queries into analytical statements that simplify the creation of SQL statements, leading to the generation of the final SQL query. \\
    \\
    Example 1:\\
    \#\# Tables:\\
    Table department, columns = [*,Department\_ID,Name,Creation,Ranking,Budget\_in\_Billions,Num\_Employees]\\
    Table head, columns = [*,head\_ID,name,born\_state,age]\\
    Table management, columns = [*,department\_ID,head\_ID,temporary\_acting]\\
    \#\# Foreign\_keys:\\
    {[management.head\_ID = head.head\_ID,management.department\_ID = department.Department\_ID]}\\
    \#\# Query:\\
    List the name, born state and age of the heads of departments ordered by age.\\
    SQL query: SELECT name ,  born\_state ,  age FROM head ORDER BY age\\
    \\
    Example 2:\\
    \#\# Tables:\\
    Table department, columns = [*,Department\_ID,Name,Creation,Ranking,Budget\_in\_Billions,Num\_Employees]\\
    Table head, columns = [*,head\_ID,name,born\_state,age]\\
    Table management, columns = [*,department\_ID,head\_ID,temporary\_acting]\\
    \#\# Foreign\_keys:\\
    {[management.head\_ID = head.head\_ID,management.department\_ID = department.Department\_ID]}\\
    \#\# Query:\\
    List the creation year, name and budget of each department.\\
    SQL query: SELECT creation,  name,  budget\_in\_billions FROM department\\
    \\
    Example 3:\\
    \#\# Tables:\\
    Table race, columns = [*,Race\_ID,Name,Class,Date,Track\_ID]\\
    Table track, columns = [*,Track\_ID,Name,Location,Seating,Year\_Opened]\\
    \#\# Foreign\_keys:\\
    {[race.Track\_ID = track.Track\_ID]}\\
    \#\# Query:\\
    Show year where a track with a seating at least 5000 opened and a track with seating no more than 4000 opened.\\
    SQL query: SELECT year\_opened FROM track WHERE seating BETWEEN 4000 AND 5000\\
    \\
    Example 4:\\
    \#\# Tables:\\
    Table Available\_Policies, columns = [*,Policy\_ID,policy\_type\_code,Customer\_Phone]\\
    Table Claims, columns = [*,Claim\_ID,FNOL\_ID,Effective\_Date]\\
    Table Customers, columns = [*,Customer\_ID,Customer\_name]\\
    Table Customers\_Policies, columns = [*,Customer\_ID,Policy\_ID,Date\_Opened,Date\_Closed]\\
    Table First\_Notification\_of\_Loss, columns = [*,FNOL\_ID,Customer\_ID,Policy\_ID,Service\_ID]\\
    Table Services, columns = [*,Service\_ID,Service\_name]\\
    Table Settlements, columns = [*,Settlement\_ID,Claim\_ID,Effective\_Date,Settlement\_Amount]\\
    \#\# Foreign\_keys:\\
    {[Customers\_Policies.Policy\_ID = Available\_Policies.Policy\_ID,Customers\_Policies.Customer\_ID =}\\
    {Customers.Customer\_ID,First\_Notification\_of\_Loss.Customer\_ID = }\\
    {Customers\_Policies.Customer\_ID,}\\
    {First\_Notification\_of\_Loss.Policy\_ID = Customers\_Policies.Policy\_ID,}\\
    {First\_Notification\_of\_Loss.Service\_ID = Services.Service\_ID,}\\
    {Claims.FNOL\_ID = First\_Notification\_of\_Loss.FNOL\_ID, Settlements.Claim\_ID = Claims.Claim\_ID]}\\
    \#\# Query:\\
    Which policy type has the most records in the database?\\
    SQL query: SELECT policy\_type\_code FROM available\_policies GROUP BY policy\_type\_code ORDER BY count(*) DESC LIMIT 1\\
    \hline
    \caption{The prompt used for generating targeted drilling bank shots under \textbf{other simple problem}.}\label{tab:simple_prompt}
    \end{xltabular}

\subsection{Targeted Drilling Bank Auto-construction on BIRD}

In this section, we provide the specific shots generation prompt for three types of problems on the BIRD dataset.

\begin{xltabular}{1.0\linewidth}{X}
    \hline 
    \multicolumn{1}{c}{\textbf{Shots Generation Prompt of Filtering Problem}} \\
    \hline
    You are a powerful text-to-SQL reasoner. Currently, I am seeking to transform intricate text queries into analytical statements that simplify the creation of SQL statements, leading to the generation of the final SQL query. Our current focus lies in the category of filtering problems. Please learn from the provided examples, design a detailed plan for the text query, and present the resulting SQL query. \\
    \\
    Example 1:\\
    \#\# Tables:\\
    Table frpm, columns = [*,CDSCode,Academic Year,County Code,District Code,School Code,County Name,District Name,School Name,District Type,School Type,Educational Option Type,NSLP Provision Status,Charter School (Y/N),Charter School Number,Charter Funding Type,IRC,Low Grade,High Grade,Enrollment (K-12),Free Meal Count (K-12),Percent (\%) Eligible Free (K-12),FRPM Count (K-12),Percent (\%) Eligible FRPM (K-12),Enrollment (Ages 5-17),Free Meal Count (Ages 5-17),Percent (\%) Eligible Free (Ages 5-17),FRPM Count (Ages 5-17),Percent (\%) Eligible FRPM (Ages 5-17),2013-14 CALPADS Fall 1 Certification Status]\\
    Table satscores, columns = [*,cds,rtype,sname,dname,cname,enroll12,NumTstTakr,AvgScrRead,\\
    AvgScrMath,AvgScrWrite,NumGE1500]\\
    Table schools, columns = [*,CDSCode,NCESDist,NCESSchool,StatusType,County,District,School,Street,\\
    StreetAbr,City,Zip,State,MailStreet,MailStrAbr,MailCity,MailZip,MailState,Phone,Ext,Website,OpenDate,\\
    ClosedDate,Charter,CharterNum,FundingType,DOC,DOCType,SOC,SOCType,EdOpsCode,EdOpsName,\\
    EILCode,EILName,GSoffered,GSserved,Virtual,Magnet,Latitude,Longitude,AdmFName1,AdmLName1,\\
    AdmEmail1,AdmFName2,AdmLName2,AdmEmail2,AdmFName3,AdmLName3,AdmEmail3,LastUpdate]\\
    \#\# Foreign\_keys:\\
    {[frpm.CDSCode = schools.CDSCode,satscores.cds = schools.CDSCode]}\\
    \#\# Query:\\
    Which schools served a grade span of Kindergarten to 9th grade in the county of Los Angeles, and what is its Percent (\%) Eligible FRPM (Ages 5-17)?\\
    Let's think step by step.\\
    <1> Decomposition: The query requires finding schools in Los Angeles County that serve a grade span of Kindergarten to 9th grade and calculating their Percent (\%) Eligible FRPM (Ages 5-17).\\
    <2> Schema Linking: In this step, we identify the tables and columns that should be used based on the first step and the foreign key relationships. We need to select School from table 'schools' and filter by County and GSserved columns. Then, we need to calculate the Percent (\%) Eligible FRPM (Ages 5-17) using the FRPM Count (Ages 5-17) and Enrollment (Ages 5-17) columns from the 'frpm' table.\\
    <3> SQL Generation: We can use an INNER JOIN to combine the 'frpm' and 'schools' tables on the 'CDSCode' column. Then, we can filter the rows where the County is 'Los Angeles' and GSserved is 'K-9'. Finally, we can calculate the Percent (\%) Eligible FRPM (Ages 5-17) by dividing the FRPM Count (Ages 5-17) by the Enrollment (Ages 5-17) and multiplying by 100.\\
    SQL query: SELECT T2.School, T1.`FRPM Count (Ages 5-17)` * 100 / T1.`Enrollment (Ages 5-17)` FROM frpm AS T1 INNER JOIN schools AS T2 ON T1.CDSCode = T2.CDSCode WHERE T2.County = 'Los Angeles' AND T2.GSserved = 'K-9'\\
    \\
    Example 2:\\
    \#\# Tables:\\
    Table account, columns = [*,account\_id,district\_id,frequency,date]\\
Table card, columns = [*,card\_id,disp\_id,type,issued]\\
Table client, columns = [*,client\_id,gender,birth\_date,district\_id]\\
Table disp, columns = [*,disp\_id,client\_id,account\_id,type]\\
Table district, columns = [*,district\_id,A2,A3,A4,A5,A6,A7,A8,A9,A10,A11,A12,A13,A14,A15,A16]\\
Table loan, columns = [*,loan\_id,account\_id,date,amount,duration,payments,status]\\
Table order, columns = [*,order\_id,account\_id,bank\_to,account\_to,amount,k\_symbol]\\
Table trans, columns = [*,trans\_id,account\_id,date,type,operation,amount,balance,k\_symbol,bank,account]\\
    \#\# Foreign\_keys:\\
    {[account.district\_id = district.district\_id,card.disp\_id = disp.disp\_id,client.district\_id = district.district\_id,disp.client\_id = client.client\_id,disp.account\_id = account.account\_id,loan.account\_id = account.account\_id,order.account\_id = account.account\_id,trans.account\_id = account.account\_id]}\\
    \#\# Query:\\
    For loan amount less than USD100,000, what is the percentage of accounts that is still running with no issue.\\
    Let's think step by step.\\
    <1> Decomposition: The query requires finding the percentage of accounts with loans less than USD 100,000 that are still running with no issue.\\
    <2> Schema Linking: In this step, we identify the tables and columns that should be used based on the first step and the foreign key relationships. We need to filter the 'loan' table by the 'amount' column and then calculate the percentage of accounts with a 'status' of 'C', which means running with no issues.\\
    <3> SQL Generation: We can use the SELECT statement to filter the 'loan' table by the 'amount' column, where the amount is less than 100,000. Then, we can calculate the percentage of accounts with a 'status' of 'C' by dividing the number of accounts with status 'C' by the total number of accounts and multiplying by 100.\\
    SQL query: SELECT CAST(SUM(status = 'C') AS REAL) * 100 / COUNT(amount) FROM loan WHERE amount < 100000\\
    \\
    Example 3:\\
    \#\# Tables:\\
    Table atom, columns = [*,atom\_id,molecule\_id,element]\\
Table bond, columns = [*,bond\_id,molecule\_id,bond\_type]\\
Table connected, columns = [*,atom\_id,atom\_id2,bond\_id]\\
Table molecule, columns = [*,molecule\_id,label]\\
    \#\# Foreign\_keys:\\
    {[atom.molecule\_id = molecule.molecule\_id,bond.molecule\_id = molecule.molecule\_id,connected.bond\_id = bond.bond\_id,connected.atom\_id2 = atom.atom\_id,connected.atom\_id = atom.atom\_id]}\\
    \#\# Query:\\
    What is the percentage of carcinogenic molecules in triple-type bonds?\\
    Let's think step by step.\\
    <1> Decomposition: The query requires finding the percentage of carcinogenic molecules (indicated by '+') in triple-type bonds (indicated by '\#').\\
    <2> Schema Linking: In this step, we identify the tables and columns that should be used based on the first step and the foreign key relationships. We need to select the 'label' column from the 'molecule' table and the 'bond\_type' column from the 'bond' table. We also need to use the 'molecule\_id' column from the 'atom', 'molecule', and 'bond' tables to join these tables together.\\
    <3> SQL Generation: We can use an INNER JOIN to combine the 'atom', 'molecule', and 'bond' tables on the 'molecule\_id' column. Then, we can filter the rows where the bond\_type is '\#'. Finally, we can calculate the percentage of carcinogenic molecules by dividing the number of distinct carcinogenic molecules by the total number of distinct molecules and multiplying by 100.\\
    SQL query: SELECT CAST(COUNT(DISTINCT CASE WHEN T2.label = '+' THEN T2.molecule\_id ELSE NULL END) AS REAL) * 100 / COUNT(DISTINCT T2.molecule\_id) FROM atom AS T1 INNER JOIN molecule AS T2 ON T1.molecule\_id = T2.molecule\_id INNER JOIN bond AS T3 ON T2.molecule\_id = T3.molecule\_id WHERE T3.bond\_type = '\#'\\
    \hline
    \caption{The prompt used for generating targeted drilling bank shots under \textbf{filtering problem} on the BIRD dataset.}\label{tab:filter_prompt_bird}
    \end{xltabular}

\begin{xltabular}{1.0\linewidth}{X}
    \hline 
    \multicolumn{1}{c}{\textbf{Shots Generation Prompt of Combination Problem}} \\
    \hline
    You are a powerful text-to-SQL reasoner. Currently, I am seeking to transform intricate text queries into analytical statements that simplify the creation of SQL statements, leading to the generation of the final SQL query. Our current focus lies in the category of combination problems. Please learn from the provided examples, design a detailed plan for the text query, and present the resulting SQL query. \\
    \\
    Example 1:\\
    \#\# Tables:\\
    Table badges, columns = [*,Id,UserId,Name,Date]\\
Table comments, columns = [*,Id,PostId,Score,Text,CreationDate,UserId,UserDisplayName]\\
Table postHistory, columns = [*,Id,PostHistoryTypeId,PostId,RevisionGUID,CreationDate,UserId,Text,\\
Comment,UserDisplayName]\\
Table postLinks, columns = [*,Id,CreationDate,PostId,RelatedPostId,LinkTypeId]\\
Table posts, columns = [*,Id,PostTypeId,AcceptedAnswerId,CreaionDate,Score,ViewCount,Body,\\
OwnerUserId,LasActivityDate,Title,Tags,AnswerCount,CommentCount,FavoriteCount,LastEditorUserId,\\
LastEditDate,CommunityOwnedDate,ParentId,ClosedDate,OwnerDisplayName,LastEditorDisplayName]\\
Table tags, columns = [*,Id,TagName,Count,ExcerptPostId,WikiPostId]\\
Table users, columns = [*,Id,Reputation,CreationDate,DisplayName,LastAccessDate,WebsiteUrl,Location,\\
AboutMe,Views,UpVotes,DownVotes,AccountId,Age,ProfileImageUrl]\\
Table votes, columns = [*,Id,PostId,VoteTypeId,CreationDate,UserId,BountyAmount]\\
    \#\# Foreign\_keys:\\
    {[badges.UserId = users.Id,comments.UserId = users.Id,comments.PostId = posts.Id,postHistory.UserId = users.Id,postHistory.PostId = posts.Id,postLinks.RelatedPostId = posts.Id,postLinks.PostId = posts.Id,posts.ParentId = posts.Id,posts.OwnerUserId = users.Id,posts.LastEditorUserId = users.Id,tags.ExcerptPostId = posts.Id,votes.UserId = users.Id,votes.PostId = posts.Id]}\\
    \#\# Query:\\
    Which is the most valuable post in 2010? Please give its id and the owner's display name.\\
    Let's think step by step.\\
    Firstly, the query requires the most valuable post, and the value is related to FavoriteCount column of table 'posts', so we should apply order by to it.\\
    Secondly, we need to retrieve the ids and owner's display name of posts selected from first step.\\
    Finally, based on the above analysis and requirements in user query, we only need to use tables 'users' and 'posts'.\\
    SQL query: SELECT T2.OwnerUserId, T1.DisplayName FROM users AS T1 INNER JOIN posts AS T2 ON T1.Id = T2.OwnerUserId WHERE STRFTIME('\%Y', T1.CreationDate) = '2010' ORDER BY T2.FavoriteCount DESC LIMIT 1\\
    \\
    Example 2:\\
    \#\# Tables:\\
    Table customers, columns = [*,CustomerID,Segment,Currency]\\
Table gasstations, columns = [*,GasStationID,ChainID,Country,Segment]\\
Table products, columns = [*,ProductID,Description]\\
Table transactions\_1k, columns = [*,TransactionID,Date,Time,CustomerID,CardID,GasStationID,ProductID,\\
Amount,Price]\\
Table yearmonth, columns = [*,CustomerID,Date,Consumption]\\
    \#\# Foreign\_keys:\\
    {[yearmonth.CustomerID = customers.CustomerID]}\\
    \#\# Query:\\
    Which year recorded the most consumption of gas paid in CZK?\\
    Let's think step by step.\\
    Firstly, the query requires the most consumption of gas paid in CZK, and the consumption is related to the Consumption column of table 'yearmonth'. Moreover, we need to consider the currency, which is in the table 'customers'. So, we should join these two tables based on the CustomerID.\\
    Secondly, we need to filter the records where the currency is CZK. We can do this using a WHERE clause to filter records from the 'customers' table.\\
    Thirdly, we need to group the results by year, which can be extracted from the Date column of the 'yearmonth' table. We can use the SUBSTRING function to get the year from the Date and then use GROUP BY to group the records by year.\\
    Finally, we need to order the results by the sum of consumption in descending order and select the top record to get the year with the most consumption of gas paid in CZK.\\
    SQL query: SELECT SUBSTRING(T2.Date, 1, 4) as Year FROM customers AS T1 INNER JOIN yearmonth AS T2 ON T1.CustomerID = T2.CustomerID WHERE T1.Currency = 'CZK' GROUP BY Year ORDER BY SUM(T2.Consumption) DESC LIMIT 1\\
    \\
    Example 3:\\
    \#\# Tables:\\
    Table circuits, columns = [*,circuitId,circuitRef,name,location,country,lat,lng,alt,url]\\
Table constructorResults, columns = [*,constructorResultsId,raceId,constructorId,points,status]\\
Table constructorStandings, columns = [*,constructorStandingsId,raceId,constructorId,points,position,\\
positionText,wins]\\
Table constructors, columns = [*,constructorId,constructorRef,name,nationality,url]\\
Table driverStandings, columns = [*,driverStandingsId,raceId,driverId,points,position,positionText,wins]\\
Table drivers, columns = [*,driverId,driverRef,number,code,forename,surname,dob,nationality,url]\\
Table lapTimes, columns = [*,raceId,driverId,lap,position,time,milliseconds]\\
Table pitStops, columns = [*,raceId,driverId,stop,lap,time,duration,milliseconds]\\
Table qualifying, columns = [*,qualifyId,raceId,driverId,constructorId,number,position,q1,q2,q3]\\
Table races, columns = [*,raceId,year,round,circuitId,name,date,time,url]\\
Table results, columns = [*,resultId,raceId,driverId,constructorId,number,grid,position,positionText,\\
positionOrder,points,laps,time,milliseconds,fastestLap,rank,fastestLapTime,\\
fastestLapSpeed,statusId]\\
Table seasons, columns = [*,year,url]\\
Table status, columns = [*,statusId,status]\\
    \#\# Foreign\_keys:\\
    {[races.circuitId = circuits.circuitId,races.year = seasons.year,constructorResults.constructorId = constructors.constructorId,constructorResults.raceId = races.raceId,constructorStandings.constructorId = constructors.constructorId,constructorStandings.raceId = races.raceId,driverStandings.driverId = drivers.driverId,driverStandings.raceId = races.raceId,lapTimes.driverId = drivers.driverId,lapTimes.raceId = races.raceId,pitStops.driverId = drivers.driverId,pitStops.raceId = races.raceId,qualifying.constructorId = constructors.constructorId,qualifying.driverId = drivers.driverId,qualifying.raceId = races.raceId,results.statusId = status.statusId,results.constructorId = constructors.constructorId,results.driverId = drivers.driverId,results.raceId = races.raceId]}\\
    \#\# Query:\\
    List out the top 10 Spanish drivers who were born before 1982 and have the latest lap time.\\
    Let's think step by step.\\

Firstly, the query requires the top 10 Spanish drivers born before 1982. We need to consider the drivers' nationality and date of birth, which are in the table 'drivers'. We can use a WHERE clause to filter records based on these conditions.\\

Secondly, we need to find the latest lap time for these filtered drivers. The lap time information is in the table 'pitStops'. So, we should join the 'drivers' and 'pitStops' tables based on the driverId.\\

Thirdly, we need to order the results by the lap time in descending order to get the latest lap time. We can use the ORDER BY clause for this purpose.\\

Finally, we need to select the top 10 records to get the required result.\\

SQL query: SELECT T2.driverId FROM pitStops AS T1 INNER JOIN drivers AS T2 on T1.driverId = T2.driverId WHERE T2.nationality = 'Spanish' AND STRFTIME('\%Y', T2.dob) < '1982' ORDER BY T1.time DESC LIMIT 10\\
    \hline
    \caption{The prompt used for generating targeted drilling bank shots under \textbf{combination problem} on the BIRD dataset.}\label{tab:combination_prompt_bird}
    \end{xltabular}

\subsection{QGP Prompt}
\label{app:sec:qgp_prompt}
In this section, we demonstrate our few-shot instruction prompt using in QGP sub-task~(Table~\ref{tab:qgp_prompt}).
\begin{xltabular}{1.0\linewidth}{X}
    \hline 
    \multicolumn{1}{c}{\textbf{Few-shot Prompt used in QGP sub-task}} \\
    \hline
    You are a Text-to-SQL expert. Your task is to classify text-based queries. The types are defined as follows:
1. Set operations, which require complex logical connections between multiple conditions and often involve the use of intersect, except, union, and other operations; 
2. Combination operations, which require grouping of specific objects and finding the maximum value or sorting, often achieved using GROUP BY; 
3. Filtering problems, which select targets based on specific judgment conditions, generally completed using where statements; 
4. Other simple problems, including simple retrieval and sorting.\\
Your task is to judge the query step by step to see if it belongs to a certain category. For example, if you think the query has the characteristics of the first type, then classify it as the first type without considering the subsequent types. If you think the query does not have the characteristics of the first type but has the second type, then classify it as the second type without considering the subsequent types.\\
\\
\#\# Example 1:\\
What are the ids of the students who either registered or attended a course?\\
Reason: We first consider Set operations. The query can be considered union logic which finds students that registered or attended a course, so it is classified as Set operations.\\
Type: Multi-set operations\\
\\
\#\# Example 2:\\
List the states where both the Secretary of 'Treasury' department and the Secretary of 'Homeland Security' were born.\\
Reason: We first consider Set operations. The query can be considered intersection logic which requires the intersection of states that 'Treasury' and 'Homeland Security' were born, so it is classified as Set operations.\\
Type: Multi-set operations\\
\\
\#\# Example 3:\\
Find all the zip codes in which the max dew point has never reached 70.\\
Reason: We first consider Set operations. The query can be seen as a difference logic, which removes zip codes that have reached a dew point of 70 from all zip codes, so it is classified as Set operations.\\
Type: Multi-set operations\\
\\
\#\# Example 4:\\
Find the name of customers who do not have an saving account.\\
Reason: We first consider Set operations. The query can be consiederd difference logic, which removes customers having an saving account from all customers, so it is classified as Set operations.\\
Type: Multi-set operations\\
\\
\#\# Example 5:\\
Which origin has the most number of flights?\\
Reason: We first consider Set operations. This query does not involve logical connection relationships. We secondly consider Combination operations. This query requires statistical counting of flights within different origins, so it is classified as Combination operations.\\
Type: Combination operations\\
\\
\#\# Example 6:\\
Which course is enrolled in by most students? Give me the course name.\\
Reason: We first consider Set operations. This query does not involve logical connection relationships. We secondly consider Combination operations. This query requires statistical counting of students within different courses, so it is classified as Combination operations.\\
Type: Combination operations\\
\\
\#\# Example 7:\\
Find the name of the train whose route runs through the greatest number of stations.\\
Reason: We first consider Set operations. This query does not involve logical connection relationships. We secondly consider Combination operations. This query requires statistical counting of running stations of different trains, so it is classified as Combination operations.\\
Type: Combination operations\\
\\
\#\# Example 8:\\
What are the names of musicals with nominee "Bob Fosse"?\\
Reason: We first consider Set operations. This query does not involve logical connection relationships. We secondly consider Combination operations. This query does not involve group count. We thirdly consider Filtering problems. This query needs to filter musicals based on the name of the nomenee, so it is classified as Filtering problems.\\
Type: Filtering problems\\
\\
\#\# Example 9:\\
How many distinct kinds of camera lenses are used to take photos of mountains in the country 'Ethiopia'? \\
Reason: We first consider Set operations. This query does not involve logical connection relationships. We secondly consider Combination operations. This query does not involve group count. We thirdly consider Filtering problems. This query needs to filter camera lenses based on the utilization on mountains in country 'Ethiopia', so it is classified as Filtering problems.\\
Type: Filtering problems\\
\\
\#\# Example 10:\\
How many products are there?\\
Reason: We first consider Set operations. This query does not involve logical connection relationships. We secondly consider Combination operations. This query does not involve group count. We thirdly consider Filtering problems. This query does not involve filter criteria. So it is classified as Other simple problems.\\
Type: Other simple problems\\

    \hline
    \caption{The few-shot prompt for CoT and fine-tuning used in QGP sub-task.}\label{tab:qgp_prompt}
    \end{xltabular}

\end{document}